\documentclass[10pt,conference]{IEEEtran}
\usepackage{cite}
\usepackage{amsmath,amssymb,amsfonts}
\usepackage{graphicx}
\usepackage{textcomp}
\usepackage{xcolor}
\usepackage{subfigure}
\usepackage{multirow}
\usepackage[hyphens]{url}
\usepackage{fancyhdr}
\usepackage{hyperref}
\usepackage{algorithm}
\usepackage[noend]{algpseudocode}
\usepackage{enumitem}
\usepackage{booktabs}
\usepackage{circuitikz}
\usepackage{float}

\usepackage{diagbox}

\newcommand{\thiscirc}[1]{%
\begin{circuitikz}[scale=.25, transform shape]%
\draw (0,0) to[ #1 ] (2,0); 
\end{circuitikz}%
}

\title{Algorithmic Strategies for Sustainable Reuse of Neural Network Accelerators with Permanent Faults}

\def\hpcacameraready{} 

\newcommand\hpcaauthors{Youssef A. Ait Alama$^{\dagger}$, Sampada Sakpal$^{\dagger}$, Ke Wang$^{\dagger}$, Razvan Bunescu$^{\dagger}$,\\ Avinash Karanth$^{\ddagger}$, and Ahmed Louri$^{\P}$}
\newcommand\hpcaaffiliation{University of North Carolina at Charlotte$^{\dagger}$, Ohio University$^{\ddagger}$, The George Washington University$^{\P}$\\Corresponding Author$^{\textasteriskcentered}$}
\newcommand\hpcaemail{yaitalam@charlotte.edu}

\author{
  \ifdefined\hpcacameraready
    \IEEEauthorblockN{\hpcaauthors{}}
      \IEEEauthorblockA{
        \hpcaaffiliation{} \\
        \hpcaemail{}
      }
  \else
    \IEEEauthorblockN{\normalsize{\color{white} --} \\
      \IEEEauthorblockA{
        \textcolor{white}{--} \\ 
        \textcolor{white}{--}
      }
    }
  \fi 
}

\fancypagestyle{camerareadyfirstpage}{%
  \fancyhead{}
  
  \fancyhead[C]{
    \ifdefined\aeopen
    \parbox[][12mm][t]{13.5cm}{}    
    \else
      \ifdefined\aereviewed
      \parbox[][12mm][t]{13.5cm}{}
      \else
      \ifdefined\aereproduced
      \parbox[][12mm][t]{13.5cm}{}
      \else
      \parbox[][0mm][t]{13.5cm}{}
    \fi 
    \fi 
    \fi 
    \ifdefined\aeopen 
      \includegraphics[width=12mm,height=12mm]{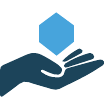}
    \fi 
    \ifdefined\aereviewed
      \includegraphics[width=12mm,height=12mm]{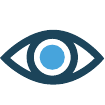}
    \fi 
    \ifdefined\aereproduced
      \includegraphics[width=12mm,height=12mm]{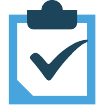}
    \fi
  }
  \fancyfoot[C]{}
}
    \makeatletter
    \newcommand{\linebreakand}{%
      \end{@IEEEauthorhalign}
      \hfill\mbox{}\par
      \mbox{}\hfill\begin{@IEEEauthorhalign}
    }
    \makeatother

\begin{document}



\maketitle 
\ifdefined\hpcacameraready 
  \thispagestyle{camerareadyfirstpage}
  \pagestyle{empty}
\else
  \thispagestyle{plain}
  \pagestyle{plain}
\fi

\newcommand{\hpcaheight}{0mm}
\ifdefined\eaopen
\renewcommand{\hpcaheight}{12mm}
\fi


\begin{abstract}
Hardware failures are a growing challenge for machine learning accelerators, many of which are based on systolic arrays. When a permanent hardware failure occurs in a systolic array, existing solutions include localizing and isolating the faulty processing element (PE), using a redundant PE for re-execution, or in some extreme cases decommissioning the entire accelerator for further investigation. In this paper, we propose novel algorithmic approaches that mitigate permanent hardware faults in neural network (NN) accelerators by uniquely integrating the behavior of the faulty component instead of bypassing it. In doing so, we aim for a more sustainable use of the accelerator where faulty hardware is neither bypassed nor discarded, instead being given a second life. We first introduce a CUDA-accelerated systolic array simulator in PyTorch, which enabled us to quantify the impact of permanent faults appearing on links connecting two PEs or in weight registers, where one bit is stuck at 0 or 1 in the {\it float32}, {\it float16}, or {\it bfloat16} representation. We then propose several algorithmic mitigation techniques for a subset of stuck-at faults, such as Invertible Scaling or Shifting of activations and weights, or fine tuning with the faulty behavior. Notably, the proposed techniques do not require any hardware modification, instead relying on existing components of widely used systolic array based accelerators, such as normalization, activation, and storage units. Extensive experimental evaluations using fully connected and convolutional NNs trained on MNIST, CIFAR-10 and ImageNet  show that the proposed fault-tolerant approach matches or gets very close to the original fault-free accuracy.
\end{abstract}

\section{Introduction and Motivation}

The substantial progress in Artificial Intelligence (AI) technology over the last two decades was made possible by the development of hardware accelerators~\cite{mishra2023artificial} together with scalable machine learning models comprised of algorithms~\cite{zinkevich_parallelized_2010,painegpu:iclr14} and architectures~\cite{vaswani_attention_2017,dosovitskiy_image_2020} that can leverage hardware-based parallelism in order to efficiently process large amounts of data.
The wide deployment of hardware accelerators for speeding up training and inference workloads in large data centers has led to an increasing number of instances where unexpected results caused either by transient or permanent failures have been reported~\cite{hochschild_cores_2021,dixit_detecting_2022,he_understanding_2023}. While significant progress has been made in understanding and characterizing the impact of hardware failures on training performance \cite{he_understanding_2023}, some of the solutions for mitigating failures are to bypass the faulty processing element (PE) in the systolic array \cite{zhang_analyzing_2018}, to re-execute using a redundant PE or even to decommission the entire TPU accelerator and resume the affected workloads on healthy devices \cite{bonderson:talk21}. In the case of permanent faults, which is the focus of this paper, a bit stuck at 0 or 1 in a link between PEs or within a TPU-internal buffer may render the component unusable in critical applications. Discarding a faulty component or bypassing it is a wasteful decision and forms part of the linear economy pattern of {\it create-use-dispose}, a manufacturing approach that is unsustainable~\cite{jawahir_technological_2016}. Disposing of a faulty component is also impractical in situations where the hardware cannot be easily replaced, for instance if it is already deployed in an artificial satellite or a space station. The downtime incurred in replacing faulty hardware can also become significant, particularly for high error rates. Therefore, it is imperative to consider feasible alternatives that still utilize the faulty component. 

In this paper, we propose novel \textbf{algorithmic methods for mitigating permanent hardware faults in systolic array based NN accelerators}. The proposed algorithms work with the faulty hardware, aiming to match the original, fault-free accuracy for stuck-at faults at particular bit positions under various floating point representations. Importantly, the proposed approach requires {\bf no additional hardware}, instead relying on already existing components in widely used systolic array accelerators, such as TPU\cite{zhang_analyzing_2018}, and SIMD accelerators, such as NVDLA~\cite{zhou2018nvdla}. We begin by characterizing the impact that stuck-at faults in links and weight registers at various locations in a systolic array have on the performance of neural networks (NN). To this end, we developed a CUDA-accelerated {\it software simulator of systolic array (S3A)}, complete with fault injection for link and weight register faults.
We then leverage the insights from fault characterization to develop fault mitigation techniques where, depending on the position of the stuck-at faulty bit, we use one of three algorithmic approaches:
\begin{itemize}
    \item {\bf Invertible Scaling and Shifting (IScSh)} is novel technique where a network's inputs, weights and activations are scaled and shifted such that the values transmitted on the affected link match the fixed value of the faulty bit and thus ensure that the fault has no impact on the network output. This effectively addresses stuck-at faults occurring in the exponent range and the sign bit (Sections~\ref{sec:isc}).
    \item {\bf Elementary Tile Operations (ETOps)} is a novel application of elementary matrix operations that strategically rearranges rows of the weight tile and columns of the activation tile, effectively addressing stuck-at faults occurring in the sign bit of weight registers (Section~\ref{sec:etops}).
    \item {\bf Fault-Aware Fine Tuning (faFT)} is a novel approach that incorporates the faulty behavior in the computation graph and backpropagates the gradient through it, fine-tuning the neural network parameters to accommodate the faulty behavior (Section~\ref{sec:ft}).
\end{itemize}
We illustrate these mitigation techniques using the IEEE 754 single precision ({\it float32}) and half precision ({\it float16}) formats, along with Google's brain floating point ({\it bfloat16}). Extensive experimental evaluations using fully connected and convolutional neural networks trained on the MNIST, CIFAR-10 and ImageNet datasets show that the proposed fault-tolerant techniques match or get very close to the original fault-free accuracy for all three floating point representations, while requiring only 17.8\% additional execution time on average.

Overall, this paper offers the first in-depth study documenting successful algorithmic strategies for recovering from single stuck-at-bit faults on links and weight registers in a weight-stationary (WS) systolic array, without changing the underlying hardware.

\section{Software Simulation of Systolic Arrays (S3A)}
\label{sec:S3A}

\begin{figure}[t]
\includegraphics[width=\columnwidth]{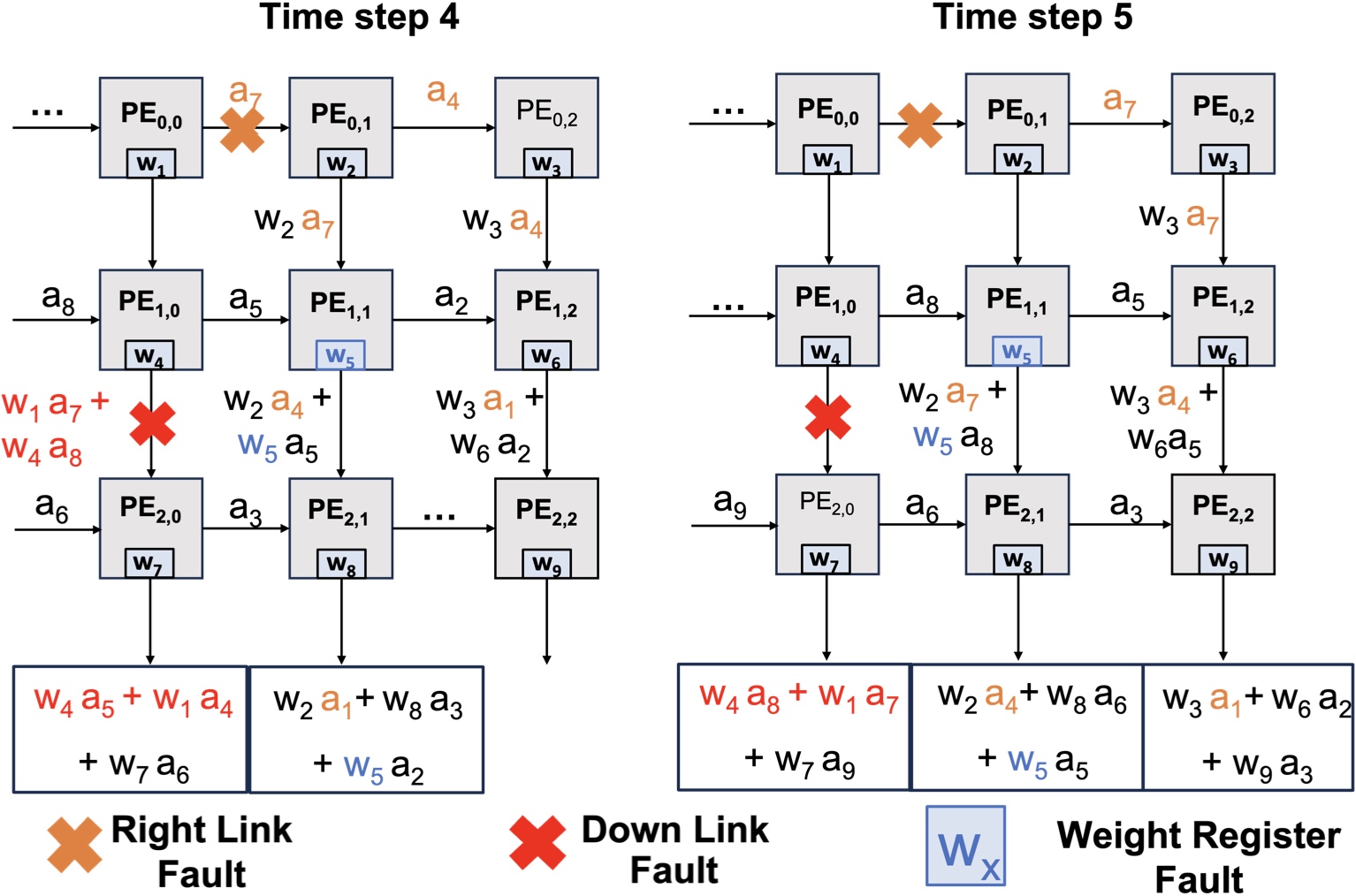}
    \caption{This figure shows snapshots in time for the fourth and fifth time step for matrix multiplication using a $3\times3$ WS systolic array with a right link fault at PE$_{0,0}$, a down link fault at PE$_{1,0}$ and a weight register fault at PE$_{1,1}$. In \textcolor{orange}{orange}, we can see the impact of the right link fault on the partial sum values that are calculated by downstream PEs. Similarly, we see the effect of down link faults in \textcolor{red}{red} and weight register faults in \textcolor{blue}{blue}.}
    \label{fig:ws_faulty_output}
\end{figure}

Systolic arrays are widely used in NN accelerators for more efficient parallel computations and data reuse~\cite{Kung}. In the context of neural networks, they can be leveraged to perform fast matrix multiplication between activation and weight matrices at each hidden layer of a neural network. By allowing a single activation value to flow through interconnected PEs, multiple partial sums are calculated efficiently in a single traversal before being passed to downstream PEs within a systolic array. The correct calculation of matrix multiplication is therefore dependent on the fault-free transfer of data passing through right and down links of a systolic array and their storage in registers within a MAC unit. A single stuck-at fault occurring in a link or register can propagate its effect to downstream PEs, as illustrated in Figure~\ref{fig:ws_faulty_output}. 

Characterizing the effect of an individual link or weight register fault in a systolic array using actual hardware can be impractical and expensive. To this aim, we developed a \href{https://github.com/yaitalam/s3a}{Software Simulation of Systolic Arrays (S3A)} in PyTorch, which also simplifies the fine-tuning NNs to incorporate the behavior of hardware faults. S3A replicates the multiply and accumulate (MAC) operations performed by PEs using PyTorch tensor operations, thus constructing a computation graph equivalent to that of a systolic array. To do this, we keep track of and update the weight, activation, and partial sum tensors at each time step. PyTorch also has the added benefit of performing automatic differentiation which allows us to easily compute gradients for fine tuning models during backpropagation. A stuck-at fault is implemented using a fault operator which modifies the value of a tensor using PyTorch tensor operations. The modified value is calculated based on the affected bits' value before and after a fault, as explained in Section~\ref{sec:ft}. The fault operator is included in the computation graph during forward propagation and backpropagation, which enables fine tuning a NN to incorporate the faulty behavior. 

It should be noted that there are also limitations to this approach. PyTorch does not guarantee bit-wise identical results for floating point operations using the IEEE 754 standard~\cite{PyTorch}. Multiplication and addition operations are not associative, so the order of operations can affect the final output even if they are mathematically identical. This loss of precision can be magnified due to frequent tensor operations used to simulate the systolic array at each time step. By comparing our results of matrix multiplication using the systolic array with PyTorch's native function, we establish a precision of up to four decimal places on average. To account for this discrepancy we compare original and fault-injection performance during inference using S3A for all experiments. Simulating a hardware component using PyTorch significantly increases the time and space complexity for storing intermediate computations of partial sums and activations at each time step. The addition of the fault operator used to inject faults at specific locations in a tensor also hinders performance. As a result, the use of this tool is limited to smaller networks and datasets for experimental evaluation.

\subsection{High-Performance S3A for fault injection using CUDA}

\begin{figure}[t]
    \centering
    \includegraphics[width=1\linewidth]{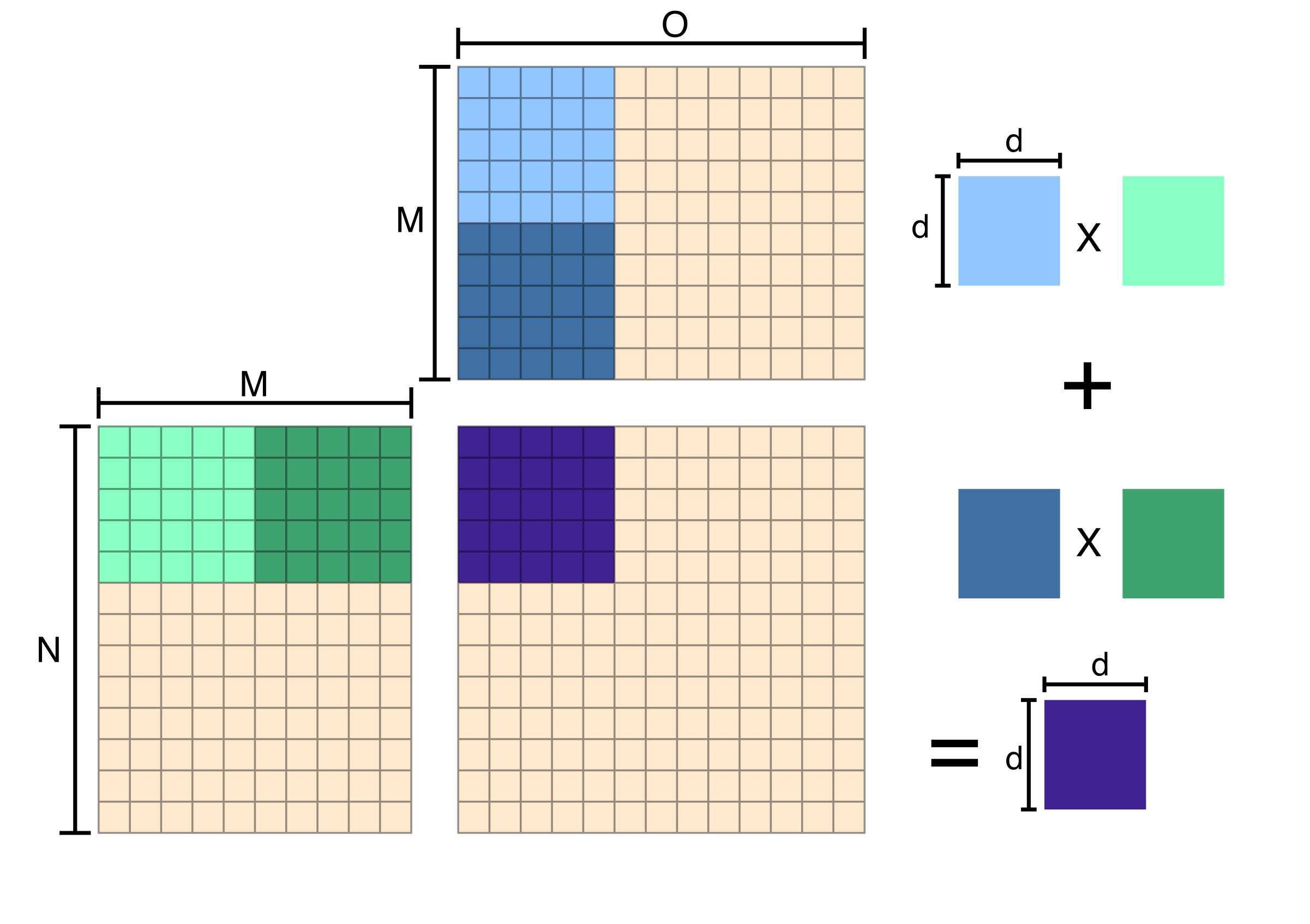}
    \vspace{-3em}
    \caption{Tiled matrix multiplication, where matrix \( A \) is \( M \times K \), matrix \( B \) is \( K \times N \), and the resulting matrix \( C \) is \( M \times N \). To tile, both matrices \( A \) and \( B \) are divided into smaller tiles of size \( K \times K \). Each tile of matrix \( A \) (in \textcolor{green}{green}) is multiplied by the corresponding tile of matrix \( B \) (in \textcolor{blue}{blue}). These multiplications produce partial products, which are then accumulated (summed) to form the final output tile in matrix \( C \) (in \textcolor{violet}{purple}).}
    \label{fig:tiledMM}
\end{figure}

This section outlines the implementation of a Systolic Array-like Matrix Multiplication (SAMM) algorithm using CUDA to enhance computational efficiency. Our approach leverages CUDA's parallel processing capabilities to accelerate this fundamental operation. To efficiently mimic faults injected within a systolic array, it is essential to understand dependencies and independencies among the various intermediate results computed by the systolic array. The core concept behind systolic array based matrix multiplication is tiling. First, the weight matrix and activation matrix are divided into smaller tiles. Then, a tile of the weight matrix is loaded into the systolic array, and a corresponding tile of the activation matrix is passed through the array from left to right, generating a partial output tile that contains their multiplication. These partial output tiles are then accumulated, as shown in Figure \ref{fig:tiledMM}. The final output tile incorporates compounded faults from each tiled multiplication step.

To leverage CUDA, each activation and weight tile pair are multiplied in parallel using the systolic multiplication procedure described in Algorithm~\ref{alg:s3a}. The computation is largely parallel, with the exception of the outer loop, which must remain sequential. This sequential step is necessary to compute faulty partial sums: partial sums for upstream PEs must be completed and passed through the fault operator before being used by downstream PEs.

\begin{algorithm}
\caption{Systolic Array MM with Fault Injection}\label{alg:s3a}
\begin{algorithmic}[1]
\Require Two matrices $A$ of size $d \times d$ and $B$ of size $d \times d$
\Require Fault position $x$, $y$, bit position $f$ and stuck value $b$
\Ensure Matrix $C$ of size $d \times d$ such that $C = A \cdot B$
    \State{$i, j \gets threadID.x, threadID.y$}
    \For{$k \gets 0$ to $d-1$}
    \State $input \gets A[i, k]$
    \State $weight \gets B[k, j]$
    \State $partial\_sum \gets C[i, j]$
        \If{fault is on right link}
            \If{$j > y$ and $k == x$}
            \State $input \gets fault\;(A[i, k] , f , b)$
            \EndIf
        \EndIf
        \If{fault type is on down link}
            \If{$j == y$ and $k == x$}
            \State $partial\_sum \gets fault\;(C[i, j] , f , b)$
            \EndIf
        \EndIf
        \If{fault is in weight register}
            \If{$j == y$ and $k == x$}
            \State $weight \gets fault\;(B[k, j] , f , b)$
            \EndIf
        \EndIf
        \State $C[i, j] \gets partial\_sum + input \cdot weight$
    \EndFor

\State \Return $C$
\end{algorithmic}
\end{algorithm}

\section{Fault Characterization}
\label{sec:characterization}

To replicate and characterize the effect of faults, we implemented four NNs of varying depths atop the S3A simulator. The smallest is a fully connected network (FCN) with two hidden layers (128 and 64 units) trained on the MNIST dataset. Moving to CNNs, the PyTorch version of LeNet \cite{lecun_gradient-based_1998} contains two convolutional layers, one max pooling layer, and three linear layers, and was trained on the CIFAR-10 dataset. Additionally, we used deeper networks, AlexNet \cite{krizhevsky_imagenet_2012} and VGG16 \cite{simonyan_very_2015} on the much larger ImageNet dataset.

\begin{figure}[t]
    \centering
    \includegraphics[width=0.95\linewidth]{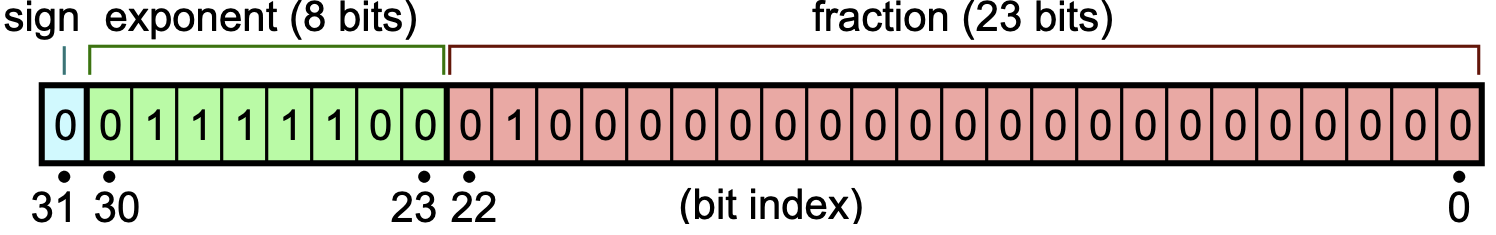}
    \caption{IEEE 754 {\it float32} representation of 0.15625.}
    \label{fig:fp-32-bit}
\end{figure}

We perform all fault characterization and mitigation using a weight stationary systolic array as shown in Figure~\ref{fig:ws_faulty_output}. In this architecture, weights are loaded into the PEs prior to feeding the input through the array. A stuck-at fault present in a link or weight register can alter the value that is transmitted to downstream PEs by setting any single bit within the floating point representation permanently to either a 0 or 1. As shown in Figure~\ref{fig:fp-32-bit} for {\it float32}, the representation consists of sign (s), exponent (e) and mantissa (m) bit ranges. Using the notation (s, e, m), we can represent the number of bits in each range for the different data types as follows: {\it float32} (1, 8, 23), {\it float16} (1, 5, 10) and {\it bfloat16} (1, 8, 7). Note that since {\it bfloat16} use the same number of bits for the exponent as {\it float32}, the two formats represent the same range of values~\cite{kalamkar2019study}.

\begin{figure*}[htp]
    \centering
    \includegraphics[width=1\linewidth]{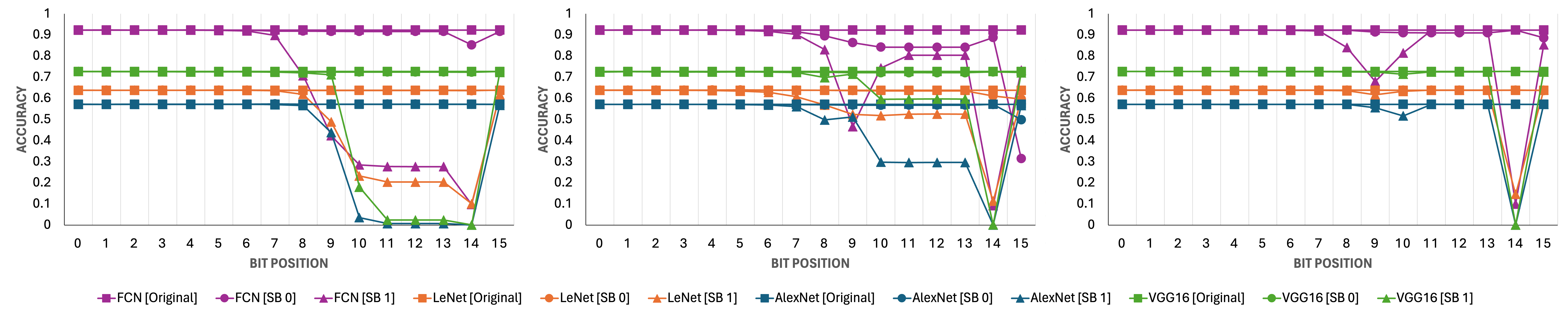}
    \caption{This figure shows the average test accuracy after a single stuck bit (SB) fault in a right link (left), down link (middle) and weight register (right) across the mantissa [0, 6], exponent [7, 14] and sign [15] bit ranges in {\it bfloat16}. Square markers denote the original fault free accuracy for a model, whereas circle and triangle markers represent the test accuracy for stuck-at-0 and stuck-at-1 faults respectively. The selected models are a FCN on the MNIST dataset, LeNet on CIFAR-10,  AlexNet and VGG16 on ImageNet.}
    \label{fig:all-models-bfloat16-fi}
\end{figure*}
\begin{figure*}[htp]
    \centering
    \includegraphics[width=1\linewidth]{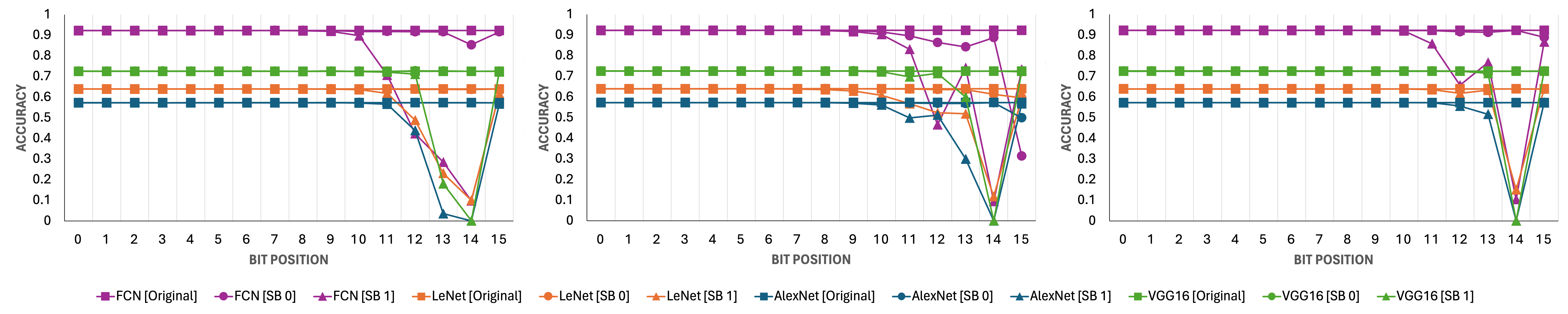}
    \caption{This figure shows the average test accuracy after a single stuck bit (SB) fault in a right link (left), down link (middle) and weight register (right) across the mantissa [0, 9], exponent [10, 14] and sign [15] bit ranges in {\tt float16}. Square markers denote the original fault free accuracy for a model, whereas circle and triangle markers represent the test accuracy for stuck-at-0 and stuck-at-1 faults respectively. The selected models are a FCN on the MNIST dataset, LeNet on CIFAR-10, AlexNet and VGG16 on ImageNet.}
    \label{fig:all-models-float16-fi}
\end{figure*}
\begin{figure*}[htp]
    \centering
    \includegraphics[width=1\linewidth]{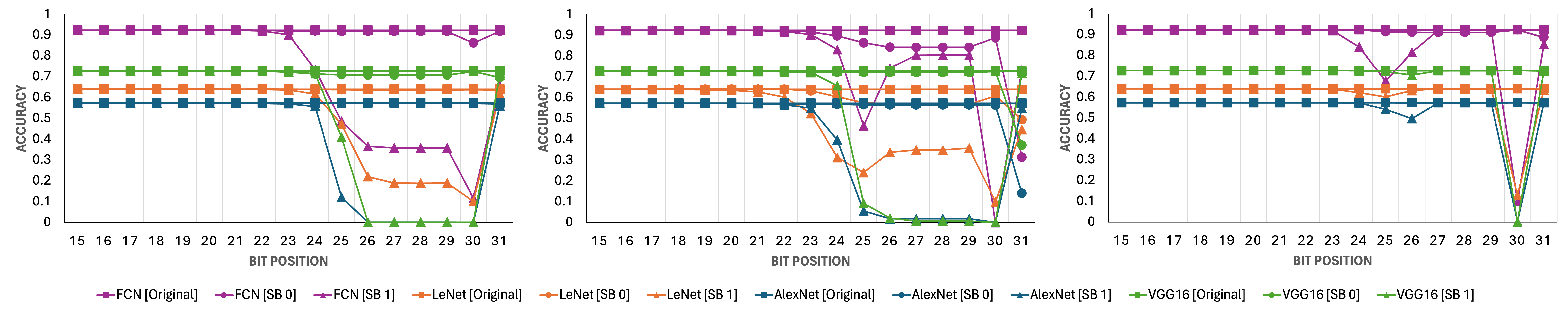}
    \caption{This figure shows the average test accuracy after a single stuck bit (SB) fault in a right link (left), down link (middle) and weight register (right) across the mantissa [0, 22], exponent [23, 30] and sign [31] bit ranges in {\tt float32}. Square markers denote the original fault free accuracy for a model, whereas circle and triangle markers represent the test accuracy for stuck-at-0 and stuck-at-1 faults respectively. The selected models are a FCN on the MNIST dataset, LeNet on CIFAR-10, AlexNet and VGG16 on ImageNet.}
    \label{fig:all-models-float32-fi}
\end{figure*}

We quantify the effect of a bit flip in each bit position by its location in the systolic array. We begin by defining and focusing on three types of stuck-at faults: {\it right link}, {\it down link} and {\it weight register} faults, as shown in Figure~\ref{fig:ws_faulty_output}. 
{\it Right Link Faults}  occur in the link connecting a PE to the one on its right in a systolic array grid. After passing through the faulty link, the faulty value persists in all partial sum calculations performed by PEs directly below and to the right of the faulty PE as shown in Figure~\ref{fig:ws_faulty_output}, thus propagating its impact throughout the systolic array. It is to be observed that a fault in a PE's input register is equivalent to a right link fault originating from the PE directly to the left of the affected PE.
{\it Down Link Faults} occur in the link connecting a PE to the one below it, or the output buffer. A faulty down link modifies the accumulated partial sum value of the affected PE, therefore, the further down a PE is within a column of the systolic array, the greater the number of accumulated partial sums affected, as shown in Figure~\ref{fig:ws_faulty_output}. It is also observed that a fault in a PE's output register is equivalent to a down link fault originating from the same affected PE.
{\it Weight Register Faults} modify the value of the preloaded weight within a PE which is used to calculate the partial sum value at any given time step. Assuming weights are still loaded correctly, the impact of this type of fault can be localized to only the partial sum contribution of the affected PE as shown in Figure~\ref{fig:ws_faulty_output}.

To observe the effect of a fault across different systolic array sizes and tiling factors, we opted to use the following \textit{dimension $\times$ PE sampling} combination from within a single row or column, depending on the nature of the fault: FCN on MNIST $\rightarrow 8 \times 8 \times$ all PEs, CNN on CIFAR-10 $\rightarrow 64 \times 64 \times$ all PEs, AlexNet on ImageNet $\rightarrow 128 \times 128 \times$ every 4th PE, and VGG16 on ImageNet $\rightarrow 256 \times 256 \times $ every 8th PE.

Figures~\ref{fig:all-models-bfloat16-fi},~\ref{fig:all-models-float16-fi}, and~\ref{fig:all-models-float32-fi} show the impact of stuck-at faults for {\it bfloat16}, {\it float16}, and {\it float32} respectively, for each of the four major NN architectures. Overall, the results indicate that the effect of a stuck-at fault largely depends on its position within the floating point format. While mantissa positions showed little to no deviation from the original accuracy after fault injection, its most significant bits still have room for improvement. Stuck-at faults in the exponent range have the greatest impact as they scale numbers exponentially from their original value, and can be deemed as some of the worst types of stuck-at faults. The effect of a sign bit can depend on factors such as the affected PE location in the systolic array and the type of link fault.
In the following we discuss in more detail the findings across the mantissa, exponent and sign bit ranges.

Mantissa stuck-at faults had minimal impact on performance across almost all of the fault types (except down links), floating point representations, models, and datasets, with an average relative error increase (REI) of less than $1\%$. Faults in reduced precision mantissa fields affected the FCN, but exhibited fault tolerance for the CNNs with smaller networks such as LeNet and AlexNet even seeing a slight improvement in accuracy for the bfloat16 format. In general, the most significant mantissa bit positions had the greatest impact on test accuracy, so we concentrate our efforts on mitigating these.

Exponent stuck-at-1 faults showed oscillating behaviour in terms of performance for down link faults. In the case of the FCN, we observe worsening performance degradation for the 3 most significant exponent bits, before a plateau until there is a sharp drop at the most significant exponent bit position. For CNNs this oscillation is dampened significantly, especially as the model increases in depth and number of parameters across all floating point representations. Deeper CNNs such as AlexNet and VGG16 experience worse performance for the most significant bit when compared to LeNet.

Exponent stuck-at-0 faults had negligible impact on performance for most cases (REI $<1\%$). We attribute this to the normalization of weights and activations before the first layer of the NN, resulting in the most significant exponent bits already being set to zero. The CNNs had a higher tolerance, especially deeper networks (AlexNet, VGG16). This may be attributed to their overparameterization that can compensate for affected neurons in lower layers so that they have less impact on the overall classification. The float32 was also more affected than float16 and bfloat16. Considering float32 and bfloat16 both share the same number of bits in the exponent field, it is possible that the reduced precision in the mantissa range of the bfloat16 format acts as a form of regularization, by restricting the range of a value that can be represented thus preventing large deviations.

Sign stuck-at faults in right links had negligible drop in performance across all models. In the case of stuck-at-0 faults, the use of the ReLU activation function in the FCN provided inherent fault tolerance by ensuring activation values passed to subsequent layers were already positive thereby negating its effect. Weight register faults also had minimal impact given their effect is limited to the partial sum contribution of the affected PE, which can potentially be compensated by healthy PEs in the affected column of systolic array. As with exponent stuck-at-0 faults, reduced precision representations appeared to be more tolerant to stuck-at sign faults, likely due to a smaller deviation from the expected fault free value.

\begin{table}[h]
\centering
\caption{Mitigation Techniques for Right Link, Down Link, and Weight Register Stuck Bit (SB) at 0 or 1 Faults Across Bit Positions}
\label{tab:mitigation_table}
\begin{tabular}{@{}ccccc@{}}

 & \multicolumn{2}{c}{\textbf{Mantissa}} & \textbf{Exponent} & \textbf{Sign}\\
\midrule
\textbf{\textit{float32} position} & 0 -- 21 & 22 & 23 -- 30 & 31 \\
\textbf{\textit{float16} position} & 0 -- 8 & 9 & 10 -- 14 & 15 \\
\textbf{\textit{bfloat16} position} & 0 -- 3 & 4 -- 6 & 7 -- 14 & 15 \\
\midrule
\textbf{Right Link SB 0} & \textit{none} & faFT & IScSh & --- \\
\textbf{Right Link SB 1} & \textit{none} & faFT & --- & --- \\
\textbf{Down Link SB 0} & \textit{none} & faFT & IScSh & IScSh \\
\textbf{Down Link SB 1} & \textit{none} & faFT & --- & IScSh \\
\textbf{Weight Register SB 0} & \textit{none} & faFT & IScSh & ETOps \\
\textbf{Weight Register SB 1} & \textit{none} & faFT & --- & ETOps \\
\bottomrule
\end{tabular}
\end{table}

\section{Algorithmic Mitigation Methods}
\label{sec:algorithm}

Built-in-self-test (BIST)~\cite{grecu2006bist} is widely used in modern computing systems to detect hardware failures, such as stuck-at faults. With the integrated test data generator and error detector, faulty links and bit positions can be easily identified. In this paper, we adopt BIST as our fault detection method. Given the stuck-at fault location and the value of the faulty bit, we propose to resolve a subset of faults by using a targeted mitigation approach, as summarized in Table~\ref{tab:mitigation_table}, where the selection of the algorithm to apply in each scenario was informed by the findings from the characterization experiments.  Stuck-at faults in the exponent range for down links or weight registers are resolved using a new {\bf Invertible Scaling and Shifting (IScSh)} technique, whereas exponent stuck-at 0 faults are addressed using only Invertible Scaling (ISc) (Section~\ref{sec:isc}). Sign bit faults in weight registers are addressed using targeted {\bf Elementary Tile Operations (ETOps)} (Section~\ref{sec:etops}), whereas mantissa stuck-at faults are mitigated using a new {\bf fault-aware Fine Tuning (faFT)} technique (Section~\ref{sec:ft}). 
It should be noted that even if all of the aforementioned fault scenarios could in theory be resolved solely by hardware modifications, in practice the cost of doing so is prohibitive. For instance, weight register faults can only be mitigated with redundant registers or PE bypassing. However, redundant registers induces substantial chip area overhead, whereas PE bypassing results in performance degradation due to unutilized PEs and additional latency penalty for PE array synchronization.

\subsection{Invertible Scaling (ISc) and Shifting (IScSh)}
\label{sec:isc}

\begin{figure*}[t]
    \centering
    \includegraphics[width=\textwidth]{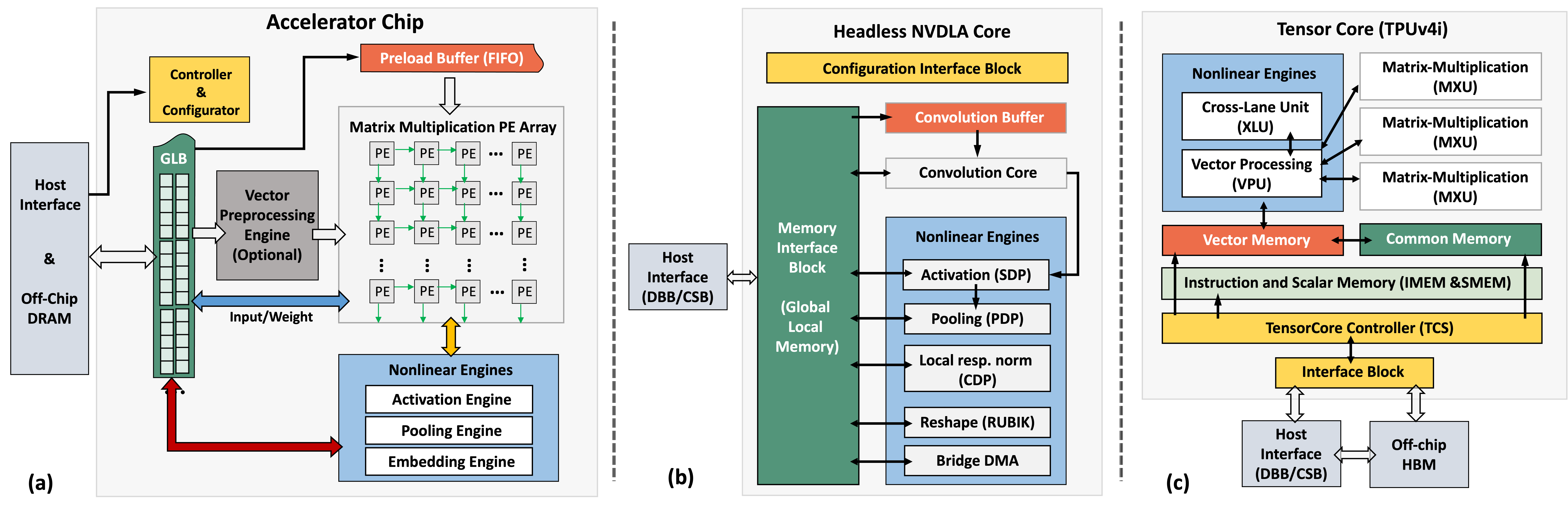}
    \vspace{-1em}
    \caption{Examples of hardware architectures that the proposed fault-tolerant algorithms can be deployed on. (a) A typical accelerator chip that shows the essential hardware components to enable the proposed  algorithms, containing a matrix multiplication array,  global and on-chip buffers, and nonlinear computation engines for activation, embedding, normalization, and others. (b) and (c) show two commercial architectures that follow the same architectural abstraction as (a): (b) shows the NVDLA~\cite{zhou2018nvdla} chip and (c) shows the TPU~\cite{zhang_analyzing_2018}. The hardware components that have similar functionalities are shown in the same colors.}
    \label{fig:architecture}
\end{figure*}

A stuck-at fault within the exponent range can cause floating point numbers to deviate drastically from their original values or stay unaffected, depending on the bit position and whether the number matches the stuck-at fault value. 
For instance, the number 1.25, represented in {\it float32} as $00\hat{1}1111110100...0$, when passed through a faulty link with exponent bit position 29 stuck at 0, changes into $00\hat{0}1111110100...0$, which corresponds to $1.25 \times 2^{-64}$. However, the number 2.25, represented in {\it float32} as $01\hat{0}0000000010...0$, when passed through the same link, would experience no change in value.
If all values passing through the faulty component align with the stuck-at value at the faulty bit position, the fault would not affect computation. This is the main idea behind invertible scaling and shifting: to mitigate stuck-at-0 faults in the exponent range of any floating point format for either right and down links or weight registers, we scale activations and weights such that the systolic array's range of values ensures all bits in the exponent range [$f$, $m$] are 0 (where $f$ is the faulty bit position and $m$ the most significant exponent bit). The NN inference algorithm using Invertible Scaling and Shifting is summarized in Table~\ref{tab:algorithm}. Each step is explained below, where we use the diode symbol \thiscirc{empty diode} to mark text explaining how calculations and storage are mapped to the generic components of the accelerators in Figure~\ref{fig:architecture}.

\begin{table}[t]
    \centering
    \caption{{Invertible Scaling (ISc) and Shifting (IScSh)} NN Inference.}
    \vspace{-1em}
    \begin{tabular}{@{}l@{}}
         \hline     
    \small $[S_1]$ Scale tiles of input $A^{(0)}$ \& weights $W^{(l)}$ into $\hat{A}^{(0)}$ \& $\hat{W}^{(l)}$ \\
    \small $[S_2]$ For each NN layer $l = 1$ to $L$:\\
    \hspace{0.2cm} \small $[S_{2.1}]$ Multiply scaled tiles of $\hat{A}^{(l-1)}$ and $\hat{W}^{(l)}$ into $\tilde{A}^{(l)}$ \\
    \hspace{0.2cm} \small $[S_{2.2}]$ Rescale activation tiles of  $\tilde{A}^{(l)}$ into scaled $\hat{A}^{(l)}$ \\
    $[S_3]$ \small Unscale tiles of $\tilde{A}^{(L)}$ to recover original NN output $A^{(L)}$ \\
         \hline
    \end{tabular}
    \label{tab:algorithm}
\end{table}
$\blacktriangleright$ Step $S_1$: {\it Scale tiles of input $A^{(0)}$ and weights $W^{(l)}$ into $\hat{A}^{(0)}$ and $\hat{W}^{(l)}$.}

Let \(A^{(0)}\) be the input matrix, $A^{(l)}$ be the activation matrix and $W^{(l)}$ the weight matrix, where $l$ denotes a layer of the neural network, for all layers $l \in \{1,2,\dots,L\}$. We first partition any \(p \times q\) activation matrix \(A^{(l)}\) into smaller, non-overlapping \(d \times d\) tiles $A^{(l)}_{i,j}$, as shown below:
\[
A^{(l)} = \begin{bmatrix}
A^{(l)}_{1,1} & A^{(l)}_{1,2} & \dots & A^{(l)}_{1,k} \\
A^{(l)}_{2,1} & A^{(l)}_{2,2} & \dots & A^{(l)}_{2,k} \\
\vdots & \vdots & \ddots & \vdots \\
A^{(l)}_{r,1} & A^{(l)}_{r,2} & \dots & A^{(l)}_{r,k}
\end{bmatrix}
\]
where $1 \leq i \leq r = \left\lceil \displaystyle\frac{p}{d} \right\rceil$  and $1 \leq j \leq k = \left\lceil \displaystyle\frac{q}{d} \right\rceil$. A similar procedure is used for partitioning $W^{(l)}$ into weight tiles $W^{(l)}_{i,j}$.

We then scale each element of the input tile $\hat{A}_{i,j}^{(0)}$ to be within the range $[-c, +c]$ and every element of the weight tile $\hat{W}_{i,j}^{(l)}$ to be within the range $[-1/d, +1/d]$. This is shown in Equation~\ref{eq:scaling1} below, where $d$ will be the dimension of the systolic array and $c$ will be calculated such that the faulty bit becomes immaterial to the matrix multiplications.
\begin{equation}
\hat{A}_{i,j}^{(0)} = \frac{c \cdot A_{i,j}^{(0)}}{\max\left|A^{(0)}_{i,j}\right|} \;\;\;\;\; \hat{W}_{i,j}^{(l)} = \frac{W_{i,j}^{(l)}}{d \cdot \max\left|W^{(l)}_{i,j}\right|}
\label{eq:scaling1}
\end{equation}

To scale the input tile $A_{i,j}^{(0)}$, we first define a function \( \max\left|M\right| \) that computes the maximum magnitude of an element of matrix $M$, as shown in Equation~\ref{eq:abs-max-activation}. In the special case where all the elements of $M$ are 0, e.g., due to padding, we set \( \max\left|M\right| = 1\) to avoid subsequent divisions by zero.
\begin{equation}
\label{eq:abs-max-activation}
   \max\left|M\right| = \begin{cases} 
     \displaystyle\max_{m \in M} |m| & \mbox{if } M \neq 0 \\
     \hfil 1 & \mbox{if } M = 0
  \end{cases}
\end{equation}
Using this equation we can calculate the scaling factor $\max\left|A_{i,j}^{(0)}\right|$ of each tile within \(A^{(0)}\). When dividing the tiles of $A^{(0)}$ by these scaling factors, all the values of $A^{(0)}$ are guaranteed to be between $[-1,1]$. The constant $c$ is determined to ensure that any value within the interval $[-c, +c]$ has zeros in all exponent bits from the most significant bit down and including to the faulty bit position, thereby insuring the impact of the stuck-at-0 fault to be inconsequential. For example, to calculate $c$ for a faulty bit position $f$ in the {\it float32} exponent field, we proceed as follows:
\begin{itemize}
    \item If the affected bit position $30 \geq f \geq 24$:
    \begin{equation}
    \label{c-bit-24-to-30}
        c = 2^{E - 127}, \;\;\;\;\;\; \mbox{where } E = \sum_{i=23}^{f - 1} 2^{i-23}
    \end{equation}
    \item If the affected bit position $p = 23$:
    \begin{equation}
    \label{c-bit-23}
    c = 2^{-126} \times \sum_{i=1}^{23}2^{-i}
    \end{equation}
\end{itemize}
To obtain the formulas for {\it float16}, the range [30, 24] and position 23 are substituted with [14, 11] and 10, whereas 127 becomes 15 and 126 becomes 14. For {\it bfloat16}, the range [30, 24] and position 23 are substituted with [14, 8] and 7.

\thiscirc{empty diode} The $\max|\cdot|$ scaling factors for all input tiles $A_{i,j}^{(0)}$ and weight tiles $W_{i,j}^{(l)}$, as well as the scaling factor $c$, are computed within the Nonlinear Engines of Fig \ref{fig:architecture} (a) (VPU in (c) or SDP in (b)) and stored within the on-chip memory (GLB or common memory) for later use. The same Nonlinear Engines are used to compute the scaled tiles $\hat{A}_{i,j}^{(0)}$ and $\hat{W}_{i,j}^{(l)}$. These, along with the dimension of the systolic array $d$, the factor $c$, and the shifting bias $b$ are stored in GLB and on-chip buffers.

$\triangleright$ {\it Compute bias $b$ for shifting.}

To mitigate stuck-at fault in the sign bit of a down link, we will need an additional shifting operation. Let $s$ be the stuck-at value of the sign bit. We compute a bias $b = (-1)^s$ and add it to the accumulator of the top PE in the column containing the down link fault. For example, if the sign bit is stuck-at-0, the bias will be $b = (-1)^0 = 1$. Adding this bias to the top PE accumulator will shift all the partial sums to be in the range $[0, 2]$, thus effectively ensuring they match the sign indicated by the stuck sign bit. Similarly, if the sign bit is stuck-at-1, the bias will be $b = -1$ and all partial sums will be shifted to $[-2, 0]$, thus matching the faulty sign bit value. The bias is to be set to zero if there is no down link sign bit fault.

$\blacktriangleright$ Step $S_{2.1}$: {\it Multiply tiles of $\hat{A}^{(l-1)}$ and $\hat{W}^{(l)}$ into $\tilde{A}^{(l)}$.}

Tiles from scaled matrix \(\hat{A}^{(l-1)}\) and scaled weight matrix \(\hat{W}^{(l)}\) are loaded into the systolic array. The shifting bias $b$ is loaded {\color{violet}in}to the top row PEs' registers for accumulation. The systolic array then computes a scaled partial output tile \(\tilde{U}^{(l)}\):
\begin{equation}
   \label{eq:unscaledpartialouttile}
   \tilde{U}^{(l)}_{i, z, j} = \hat{A}^{(l-1)}_{i, z} \cdot \hat{W}^{(l)}_{z, j} + b
\end{equation}
Each partial output tile needs to be unscaled and shifted back before accumulation, as shown in Equation~\ref{eq:unscaledpartialouttile}, to insure correct accumulations across different systolic arrays. \begin{equation}
\label{eq:partialouttile}
    U^{(l)}_{i, z, j} = (\tilde{U}^{(l)}_{i, z, j} - b) \cdot \max\left|A^{(l-1)}_{i, z}\right| \cdot \max\left|W^{(l)}_{z,j}\right| \cdot d
\end{equation}
\thiscirc{empty diode} The computations in Equation~\ref{eq:partialouttile} are done by the Nonlinear engines of Figure \ref{fig:architecture} (a) (VPU in (c) or SDP in (b)).

Once the partial output tiles \(U^{(l)}_{i,z,j}\) are computed, they are accumulated by the normal accelerator logic into the tiles of activation matrix $\tilde{A}^{(l)}$, followed by the activation function $f$, as shown below:
\begin{equation}
\tilde{A}^{(l)}_{i, j} = f\left(\sum_{z=1}^{k} U^{(l)}_{i, z, j}\right)
\label{eq:activation1}
\end{equation}

$\blacktriangleright$ Step $S_{2.2}$: {\it Rescale tiles of  $\tilde{A}^{(l)}$ into scaled $\hat{A}^{(l)}$.}

To compute the appropriately scaled activation matrix \(\hat{A}^{(l)}\) for the next NN layer, we rescale the tiles of \(\tilde{A}^{(l)}\) as shown in Equation~\ref{eq:rescale1} below.
\begin{equation}
    \hat{A}_{i,j}^{(l)} = \frac{c \cdot \tilde{A}_{i,j}^{(l)}}{\max\left|{\tilde{A}^{(l)}_{i,j}}\right|}
    \label{eq:rescale1}
\end{equation}

\thiscirc{empty diode} The $\max{|\cdot|}$ scaling factors in Equation~\ref{eq:rescale1} are computed in the activation engine in Figure~\ref{fig:architecture} at the same time with the application of the nonlinear activation function in Equation~\ref{eq:activation1}. The scaling itself is done by the Nonlinear engines in and the scaled tiles are stored in the GLB. It can be shown that the scaling factors $\max\left|{A^{(l)}_{i,j}}\right|$ needed for the next layer computations are mathematically equivalent with the scaling factors $\max\left|{\tilde{A}^{(l)}_{i,j}}\right| / c$ used above.  As such, we will use these to replace the old factors $\max\left|A^{(l-1)}_{i,j}\right|$ in the GLB. The old values $\tilde{A}^{(l-1)}$ are also replaced with the newly computed, properly scaled activation values, $\hat{A}^{(l)}$. 

Step 2 is iterated until the last layer of the neural network, skipping step 2.2 for the last iteration.

$\blacktriangleright$ Step $S_3$: {\it Unscale  $\tilde{A}^{(L)}$ to recover NN output $A^{(L)}$.}

At the end, we recover the original output matrix as $A^{(L)} = \tilde{A}^{(L)} / c$. It can be shown mathematically that this final unscaling after the transformations from Equations~\ref{eq:scaling1} to~\ref{eq:rescale1} recovers the original output matrix. \thiscirc{empty diode} This final rescaling is done in the Nonlinear engines in Figure~\ref{fig:architecture} (e.g., VPU or SDP) and stored in the GLB, then sent back to the host system.

Note that not all the IScSH computations are always required, e.g., weight scaling is not needed for right link faults, scaling activations by $c$ is not needed for down link faults, and scaling activations is not needed for weight register faults. Table~\ref{tab:requirements_table} summarizes the necessary operation for each fault type. Note that a bit stuck at 0 in weight registers only requires the weight matrices tiles to be scaled  between $[-c,c]$.
\begin{table}[ht]
\centering
\caption{Mitigation Techniques required for each fault type.\\
RL = Right Link, DL = Down Link, SB = Stuck Bit.}
\label{tab:requirements_table}
\begin{tabular}{@{}lccccc@{}}

 & Shift & Scale $W$ & Scale $W$ & Scale $A$ & Scale $A$ \\
 & bias & $[-1,1]$ & $[-\frac{1}{d},+\frac{1}{d}]$ & $[-1,1]$ & $[-c,+c]$ \\
\midrule
\textbf{RL SB 0 (Exp.)} & -- & -- & -- & \checkmark & \checkmark\\
\textbf{DL SB 0 (Exp.)} & -- & \checkmark & \checkmark & \checkmark & -- \\
\textbf{DL SB 0 (Sign)} & \checkmark & \checkmark & -- & \checkmark & -- \\
\textbf{DL SB 1 (Sign)} & \checkmark & \checkmark & -- & \checkmark & -- \\
\bottomrule
\end{tabular}
\end{table}

\subsection{Elementary Tile Operations (ETOps)}
\label{sec:etops}

The Elementary Tile Operations (ETOps) effectively correct sign bit faults in Weight registers. The integrity of the matrix multiplication results is maintained through strategic row swapping or column inversion, as follows:

$\triangleright$ \textbf{Sign Verification}: Check if the element that is to be loaded in the faulty weight register has the same sign as the detected fault. If the signs match, no action is required.

$\triangleright$ \textbf{Row Swapping}: If the signs do not match, search within the same column of the weight tile for an element (row $j$) with the same sign as the fault (row $i$). If a matching sign element is found, swap its row $i$ with the fault's row $j$, then swap columns $i$ and $j$ in the activation tile, and stop. If no matching sing element found, continue with the next step.

$\triangleright$ \textbf{Column Inversion}: Multiply each element in the column by \(-1\). Flipping the signs of all elements in the column ensures the element at position now matches the fault's sign. Once the output tile is computed, the corresponding row is inverted back by multiplying by \(-1\) to recover the original sign.

Mathematically, the above transformations produce the original output while not having to bypass the faulty PE. \thiscirc{empty diode} These additional computations are handled within the Nonlinear engines depicted in Figure~\ref{fig:architecture} (RUBIK or XLU for swapping VPU or SDP for inversion and verification). 

\subsection{Fault-Aware Fine Tuning (faFT)}
\label{sec:ft}

Some faults cannot be mitigated using invertible scaling or shifting due to the nature of the fault or the loss in numerical accuracy. In these cases, we turn to the model itself to recover the lost accuracy, by fine tuning the weights such that they accommodate the faulty behavior. 

To integrate the faulty behavior during fine tuning, we implement the stuck-at-fault as a new computational operator $fault(x | n, b)$ that takes an input value $x$ and calculates a new floating point value when bit position $n$ is stuck at binary value $b$. 
The operator is inserted in the corresponding location in the computational graph of the systolic array, enabling the gradient of the loss to backpropagate through the fault operator. In the CUDA-accelerated implementation of S3A, the fault operator $fault(x, n, b)$ directly sets the bit position $n$ to its stuck at binary value $b$ using bitwise $\sim$, $|$ and $\&$ operators:
\begin{itemize}
    \item[] $fault(x, n, 1) = x \; | \; (1 << n)$
    \item[] $fault(x, n, 0) = x \; \& \; \sim (1 << n) $
\end{itemize}
Faults are injected during the forward pass, as such the gradient computation uses the faulty activation and weight values, for which we implemented a custom backward function. Even though the fault operator itself is not continuous everywhere, in our experiments we did not observe any significant impact on the convergence of SGD training. Fault-aware fine-tuning is performed off the faulty device, as the fault would invalidate the analytic gradients. Instead, the fault is replicated during the forward pass but not during backpropagation using our CUDA accelerate S3A simulator. The resulting gradients are used to update the model's weights, allowing the model to adapt to the faulty behavior. The updated weights are then sent back to the accelerator to replace the previous weights.

\begin{figure*}[tp]
    \centering
    \includegraphics[width=0.33\linewidth]{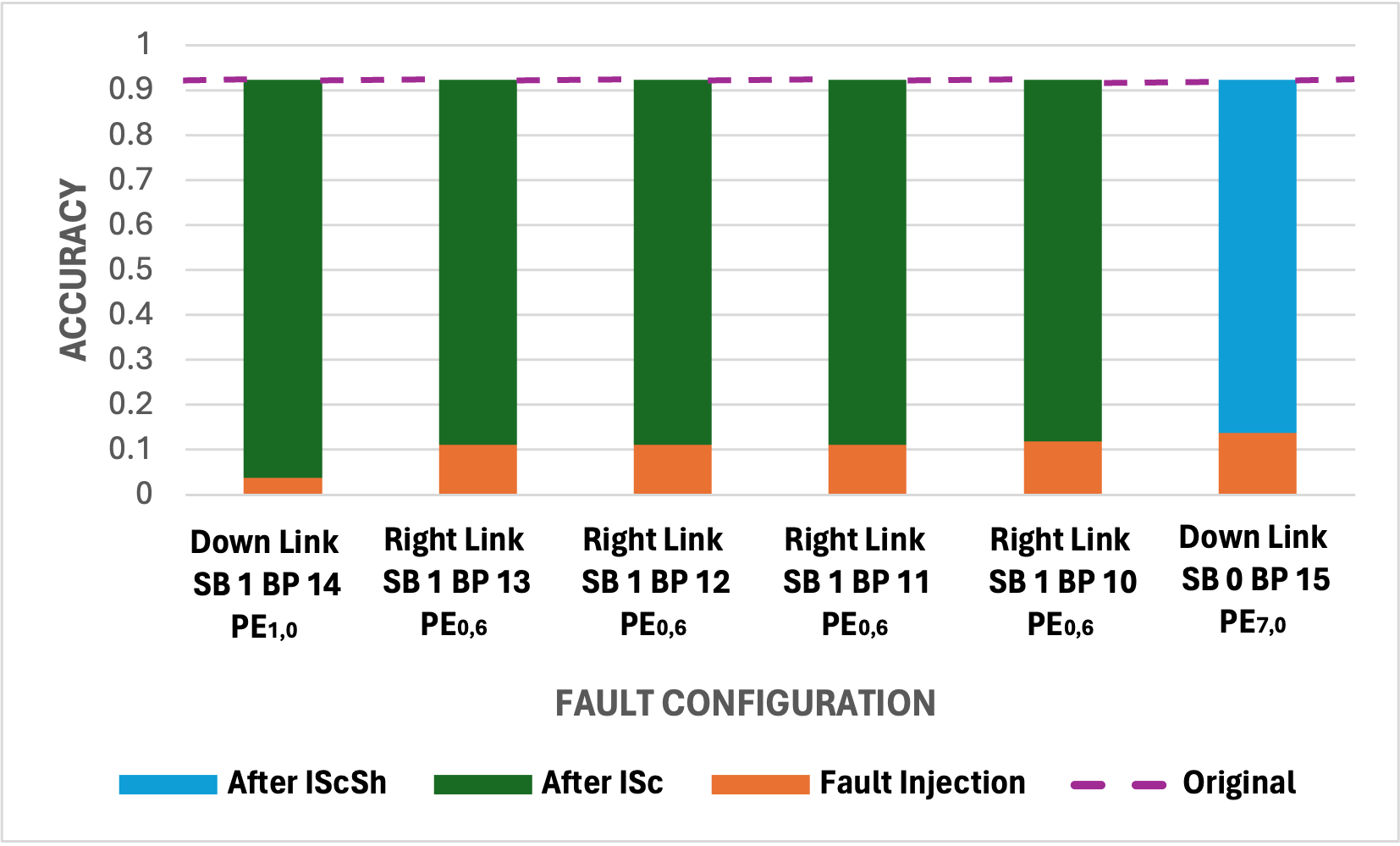}%
    \includegraphics[width=0.33\linewidth]{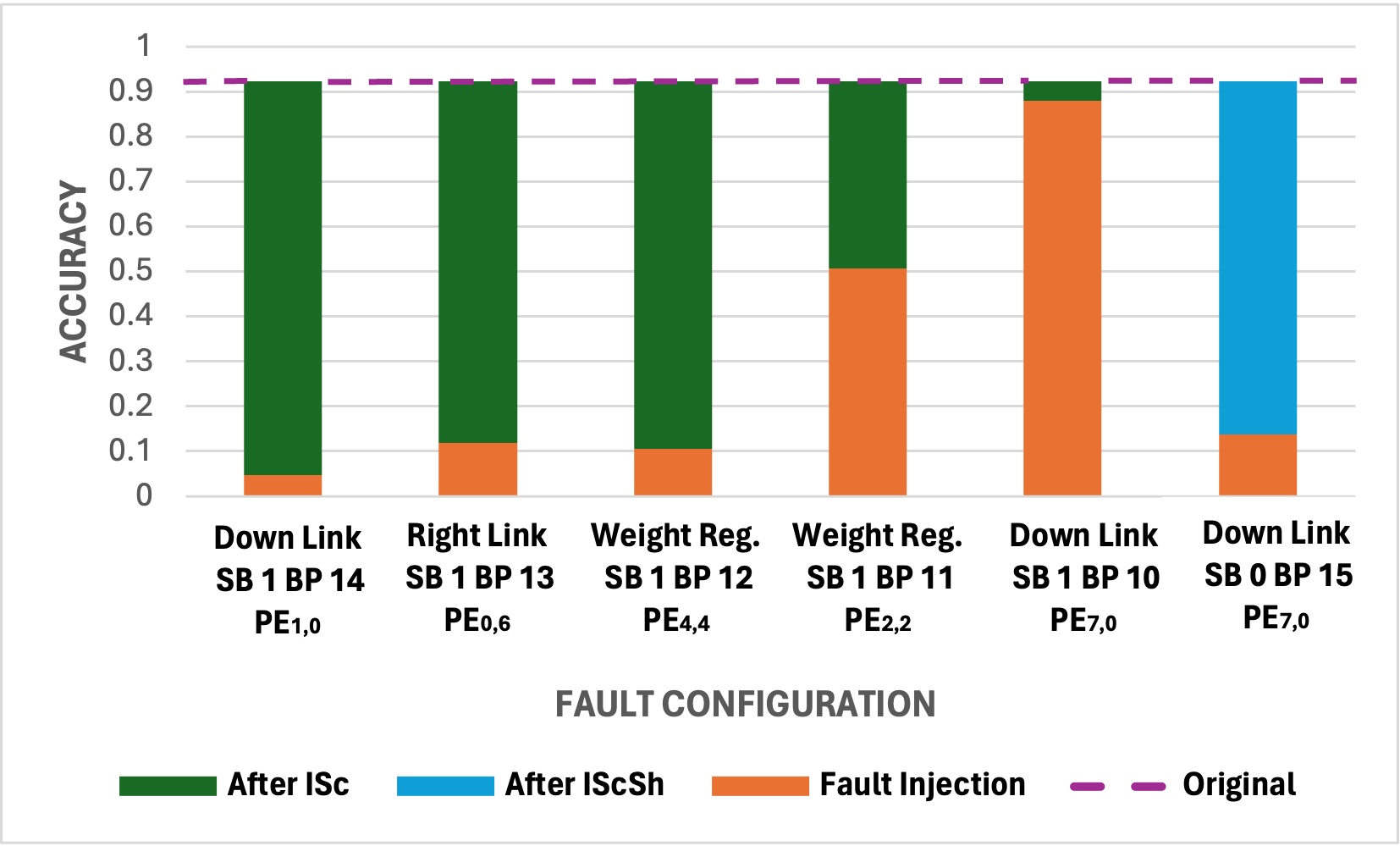}%
    \includegraphics[width=0.33\linewidth]{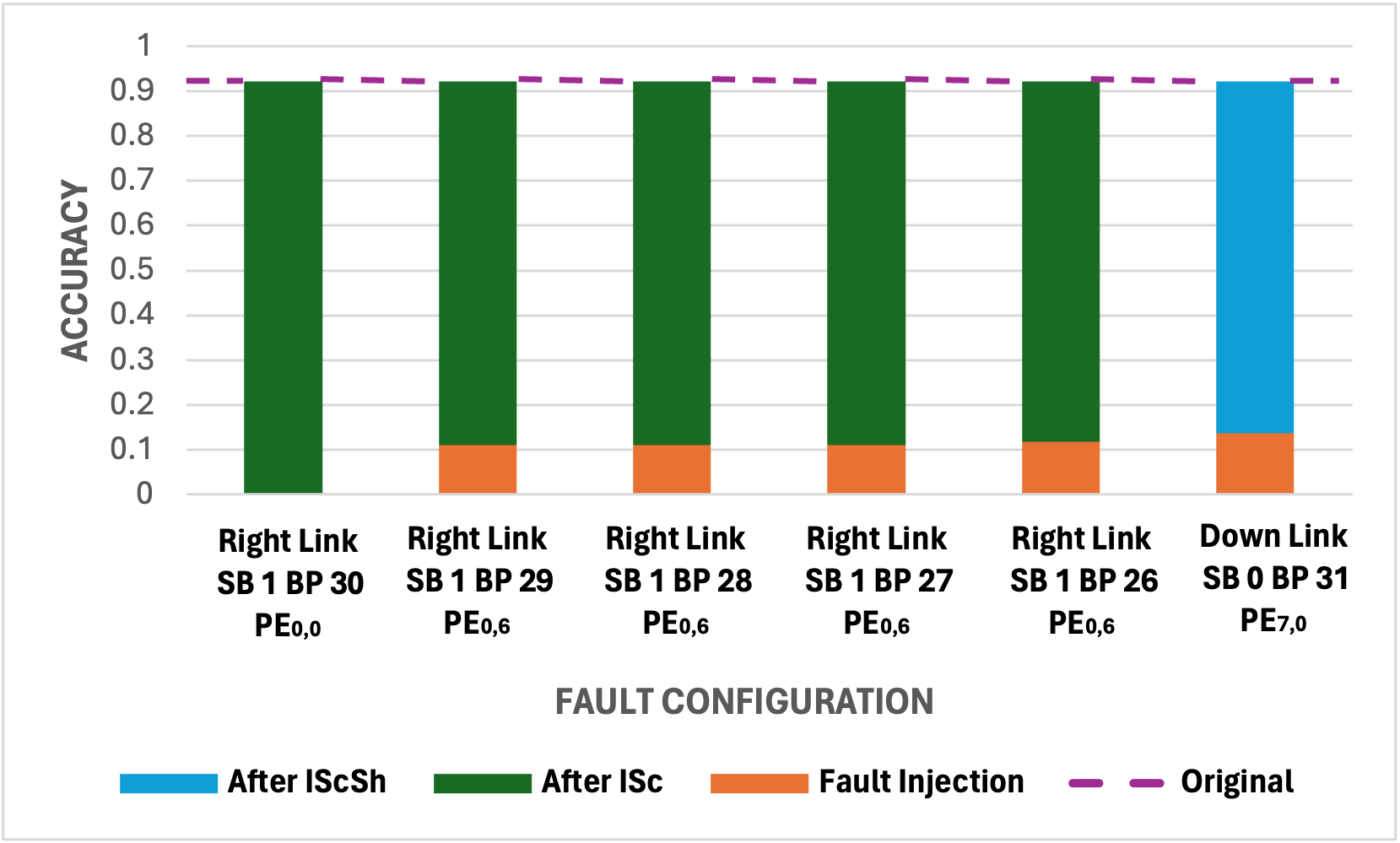}
    \caption{This figure compares the FCN performance of ISc and IScSh with the worst Fault Injection (FI) accuracies for a single stuck bit (SB) fault occurring in the exponent and sign bit positions (BP) respectively for the bfloat16 (left), float16 (middle) and float32 (right) representations. The FCN model uses the MNIST dataset, and has an original accuracy of $92.28\%$ for bfloat16 and float16, and $92.26\%$ for the float32 representation shown by the \textcolor{violet}{purple} dashed line.}
    \label{fig:fcn-iscsh}
\end{figure*}

\begin{figure*}[tp]
    \centering
    \includegraphics[width=0.33\linewidth]{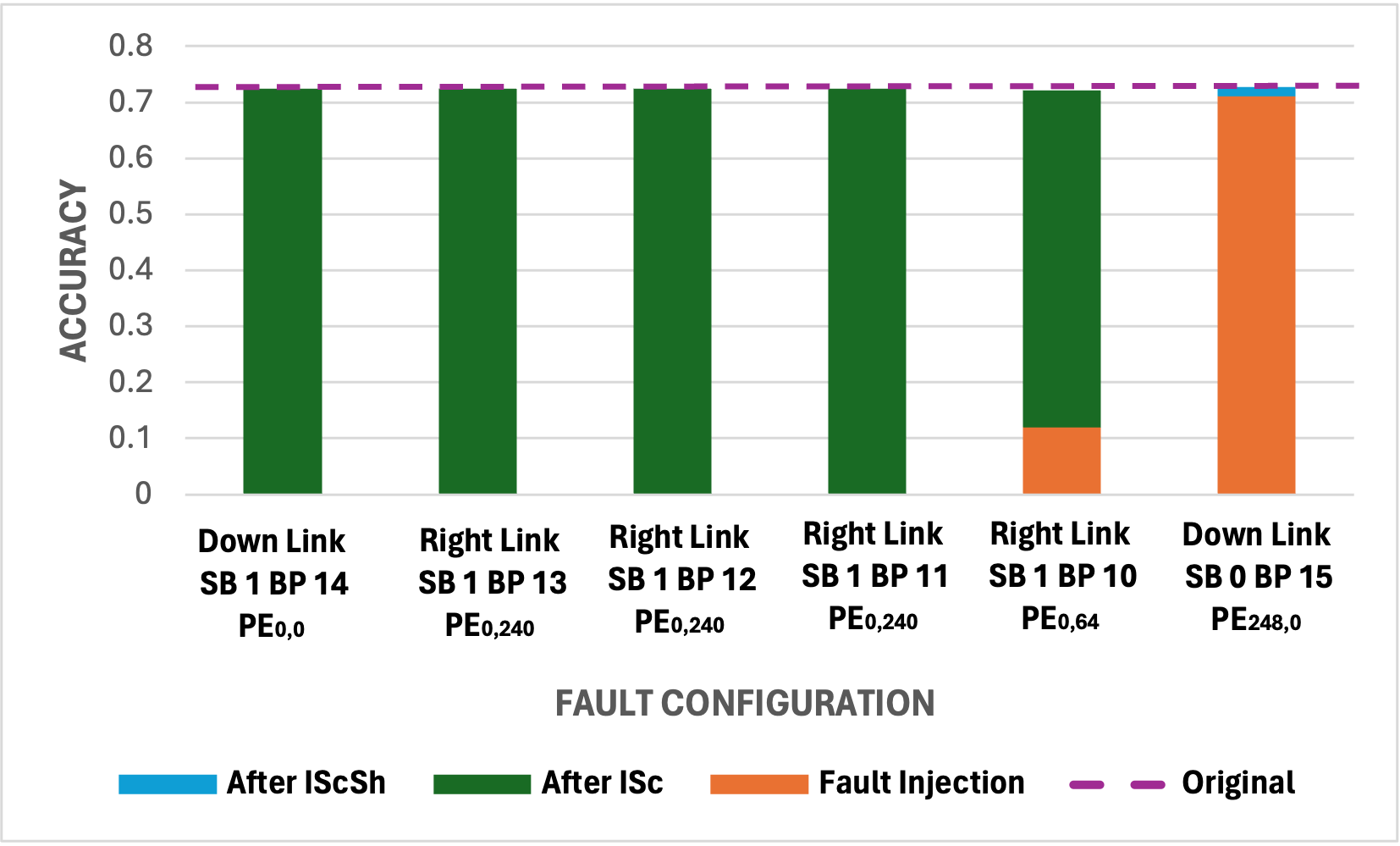}%
    \includegraphics[width=0.33\linewidth]{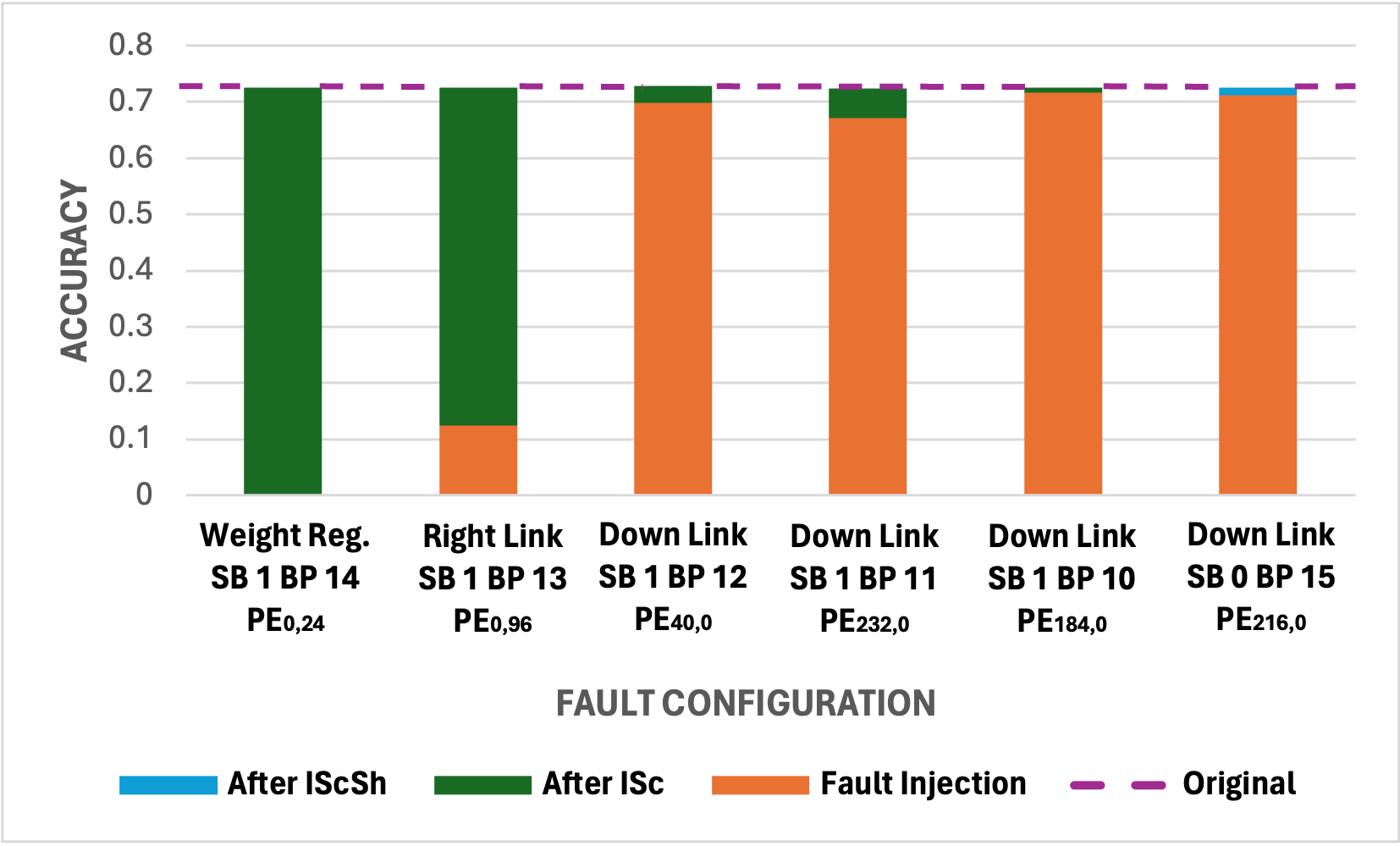}%
    \includegraphics[width=0.33\linewidth]{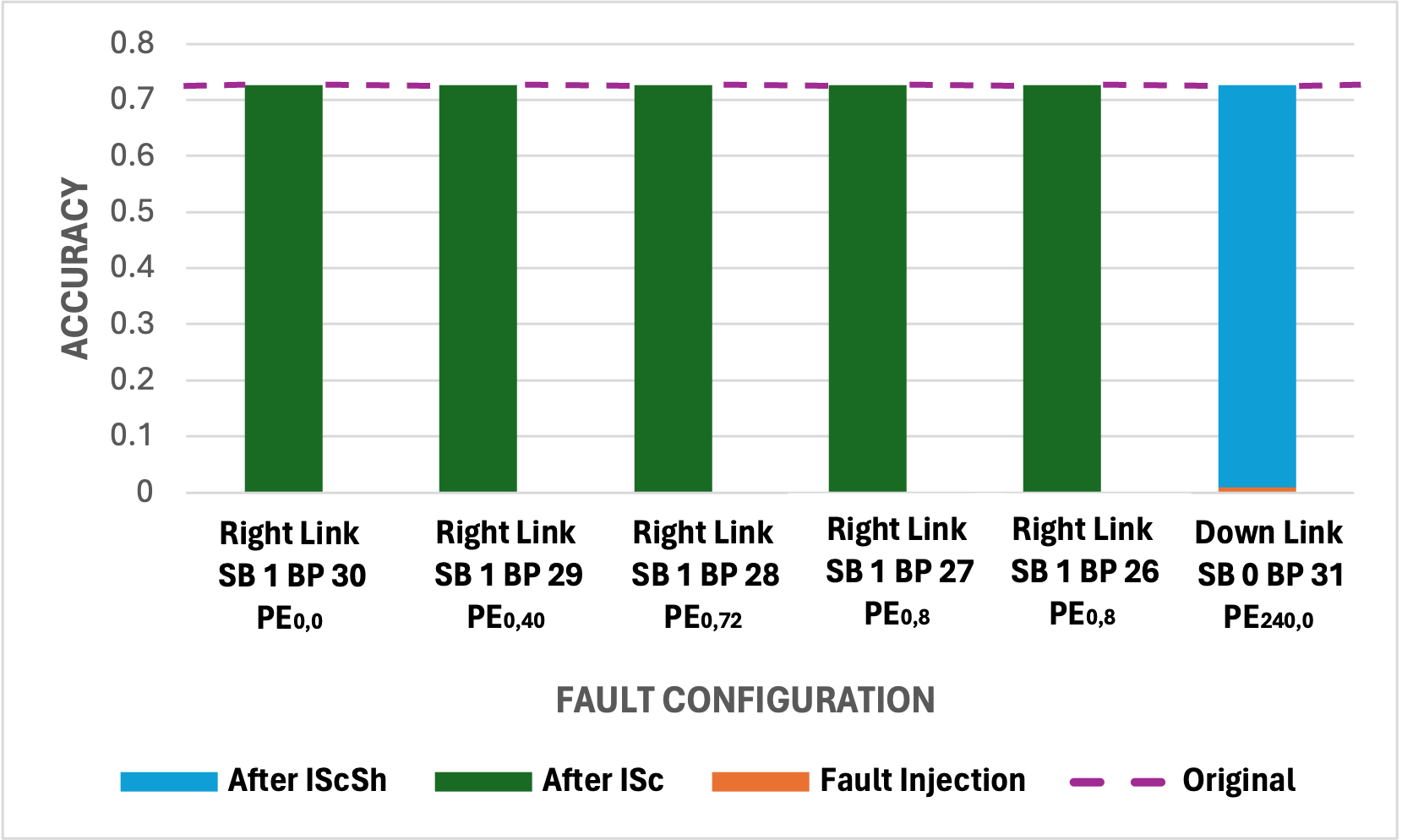}
    \caption{This figure compares the VGG16 performance of ISc and IScSh with the worst Fault Injection (FI) accuracies for a single stuck bit (SB) fault occurring in the exponent and sign bit positions (BP) respectively for the bfloat16 (left), float16 (middle) and float32 (right) data types. VGG16 uses the ImageNet dataset, and has an original accuracy of $72.7\%$ for bfloat16 and float32, and $72.5\%$ for float16 data types as shown by the \textcolor{violet}{purple} dashed line.}
    \label{fig:vgg16-iscsh}
\end{figure*}

\begin{figure*}[tp]
    \centering
    \includegraphics[width=0.33\linewidth]{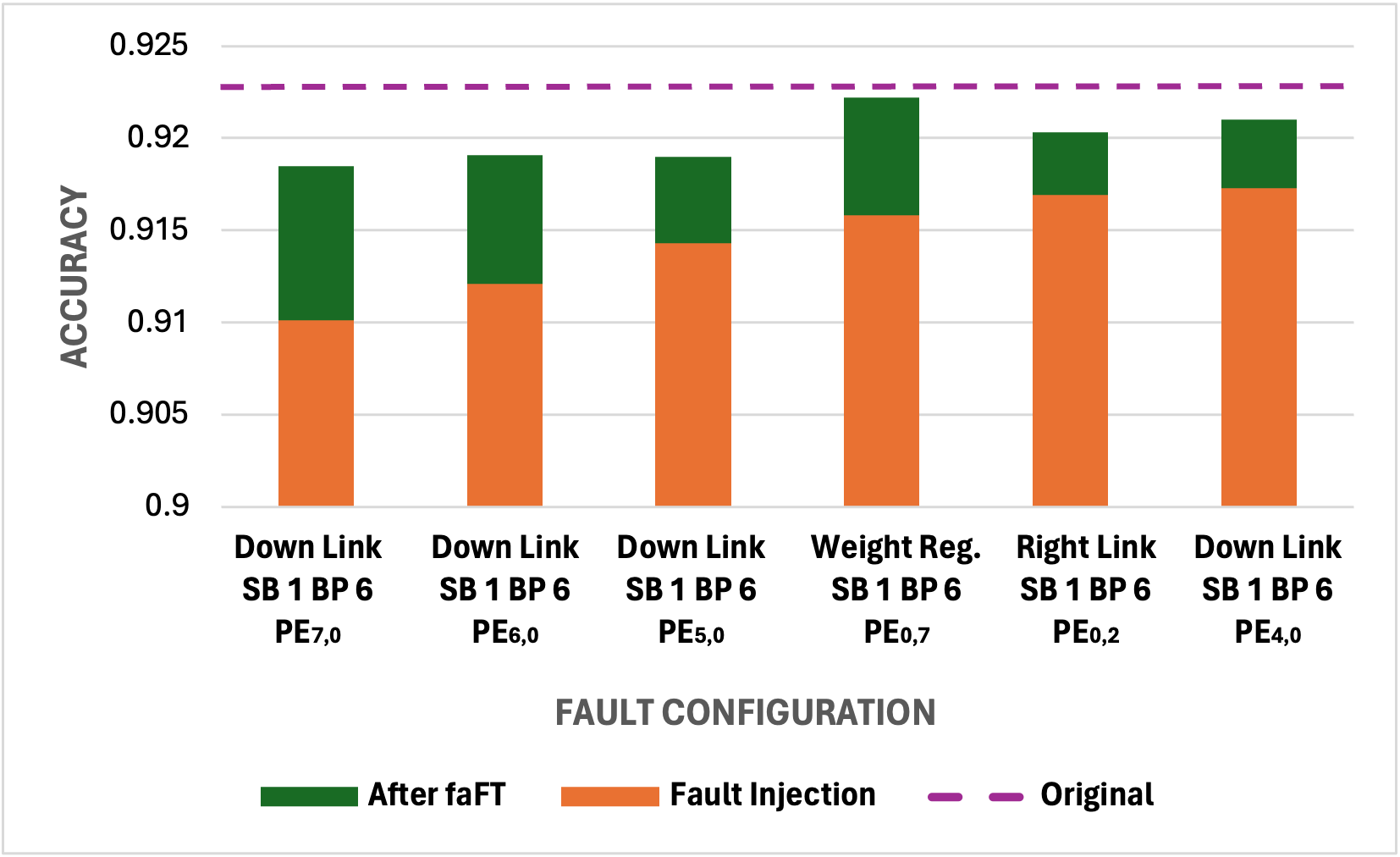}%
    \includegraphics[width=0.33\linewidth]{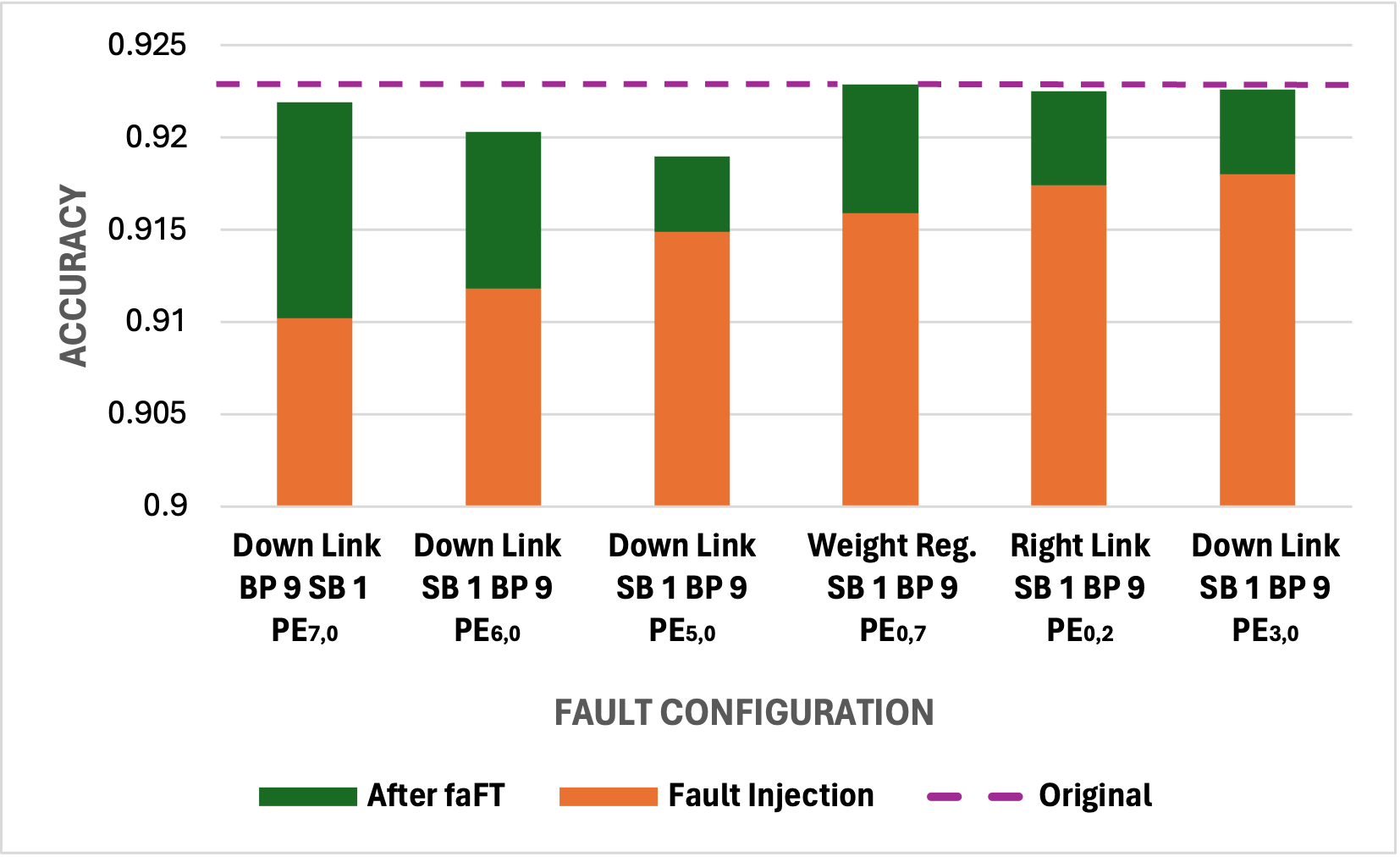}%
    \includegraphics[width=0.33\linewidth]{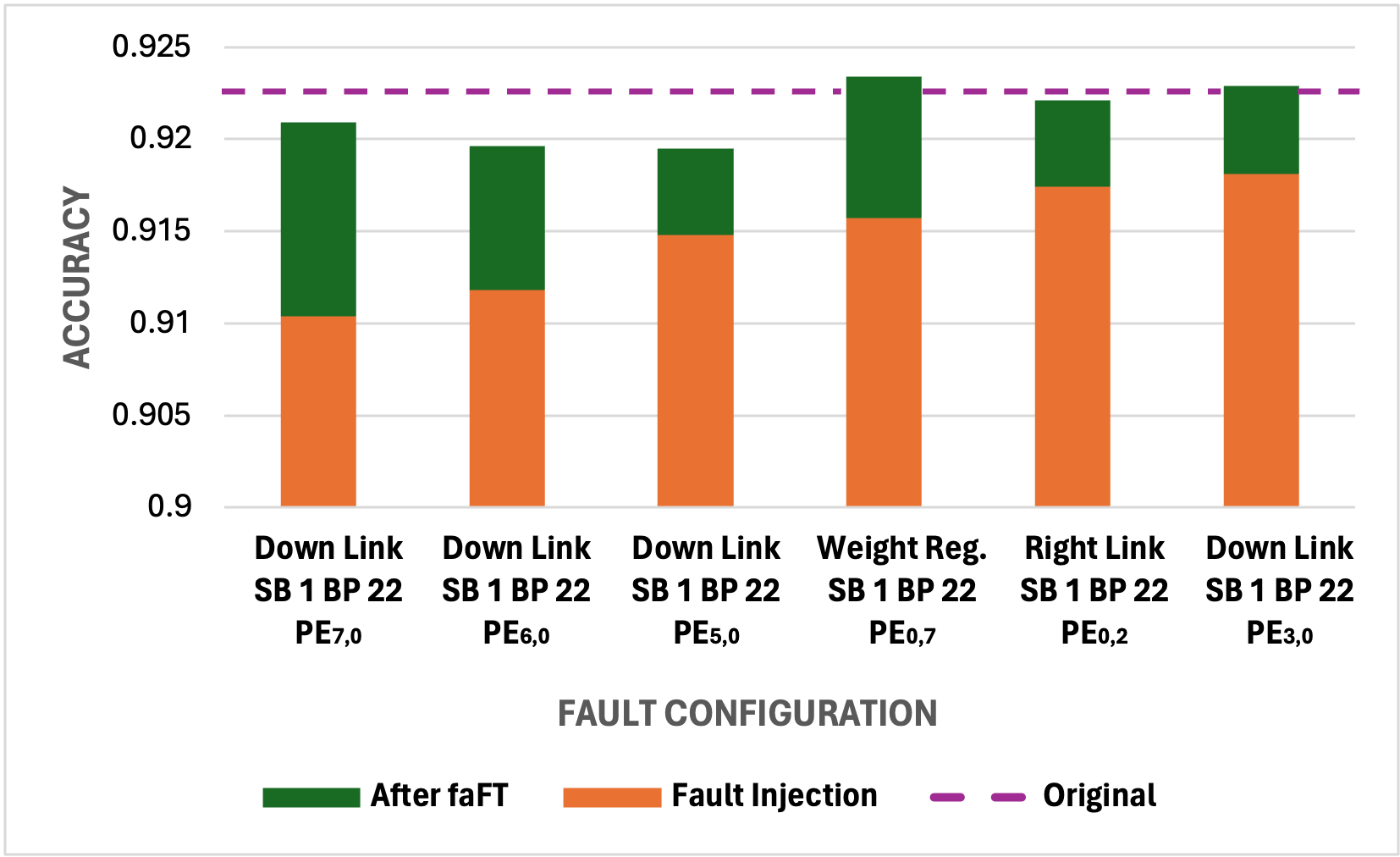}
    \caption{This figure compares the FCN performance before and after faFT for the worst affected mantissa bit position (BP) due to a stuck bit (SB) fault across the bfloat16 (left), float16 (middle) and float32 (right) representations. The original accuracy is shown by the \textcolor{violet}{purple} dashed line.}
    \label{fig:fcn-faft}
\end{figure*}

\begin{figure*}[tp]
    \centering
    \includegraphics[width=0.33\linewidth]{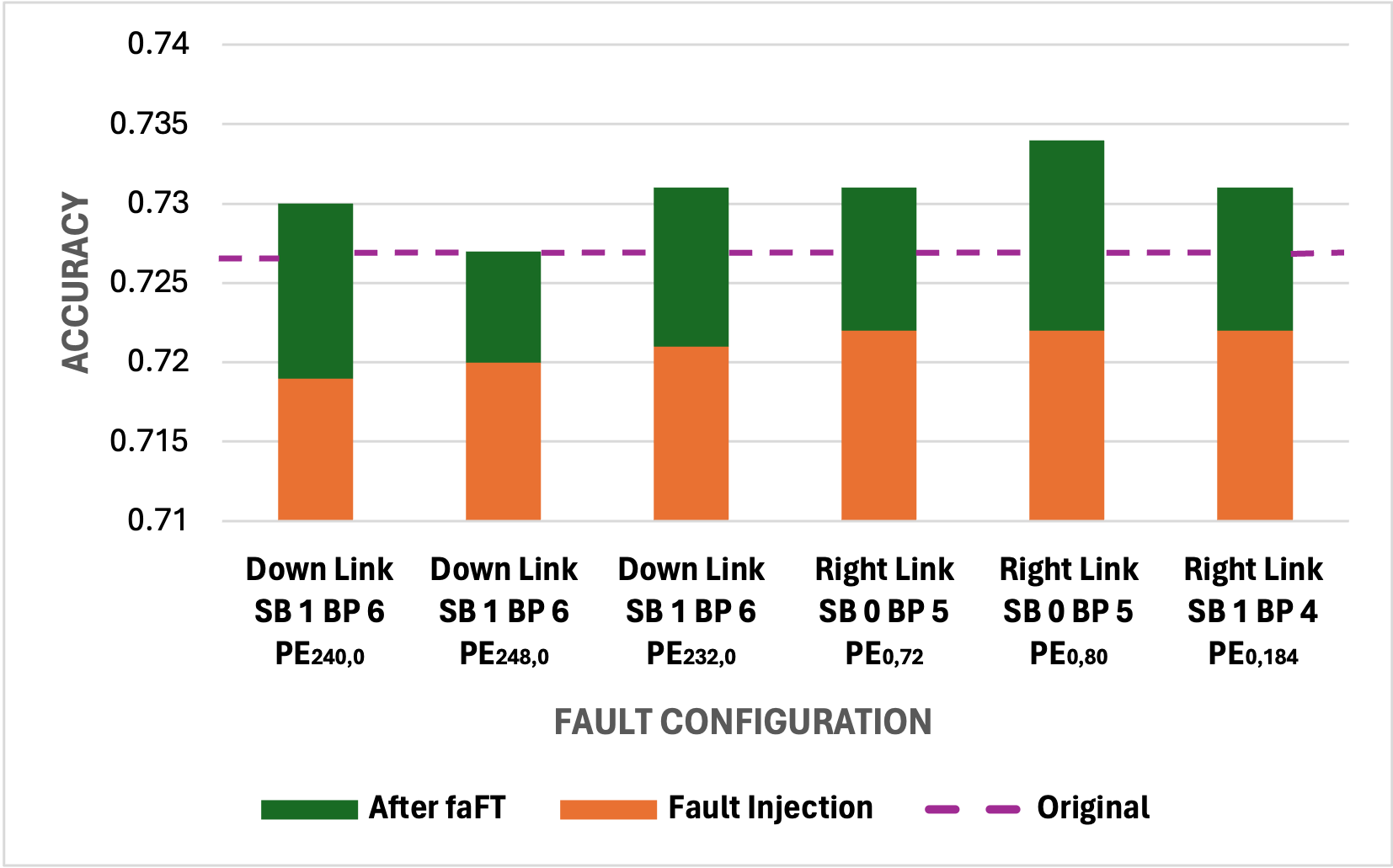}%
    \includegraphics[width=0.33\linewidth]{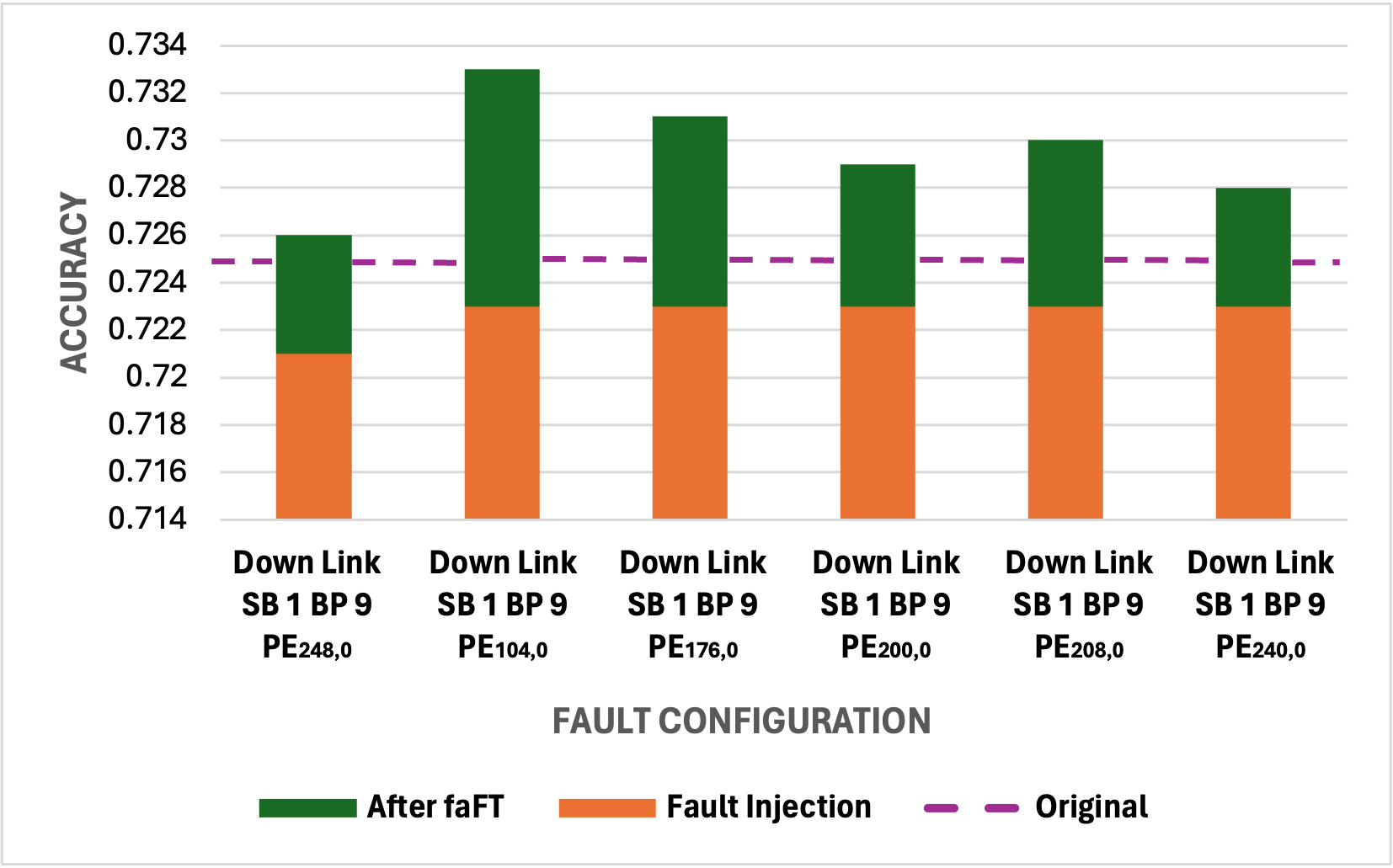}%
    \includegraphics[width=0.33\linewidth]{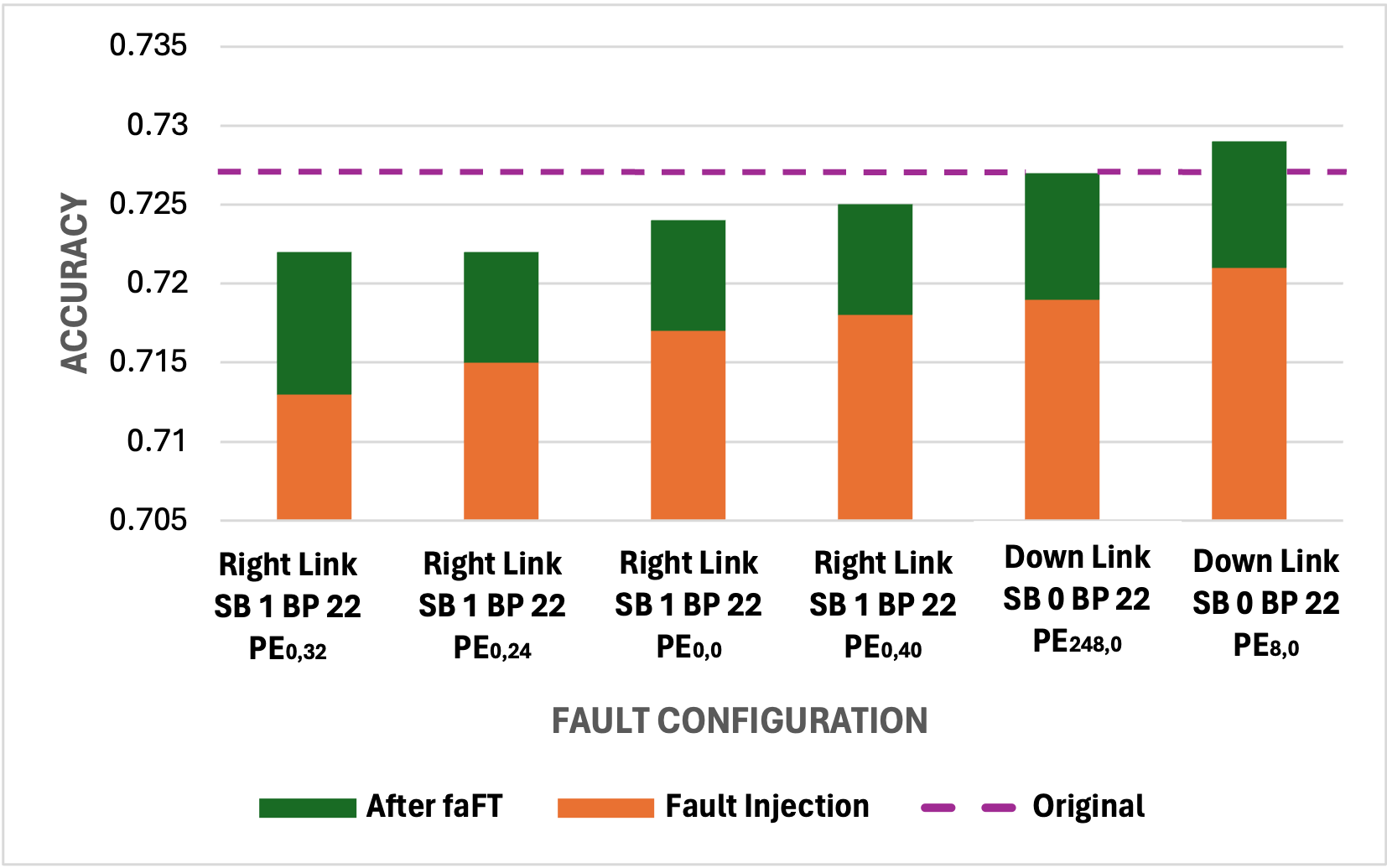}
    \caption{This figure compares the VGG16 performance of faFT for a single stuck bit (SB) fault occurring in the most significant mantissa bit positions (BP) for the bfloat16 (left), float16 (middle) and float32 (right). The original accuracy is shown by the \textcolor{violet}{purple} dashed line.}
    \label{fig:vgg16-faft}
\end{figure*}

\begin{table}[t]
    \caption{Fault-Aware Fine Tuning (faFT) for Worst Faults (Down Link Stuck-At-1) Using LeNet on CIFAR-10, AlexNet and VGG16 on ImageNet, for the most significant mantissa bit position (\%).}
    \label{tab:faft_table}
    \centering
    \begin{tabular}{llccc}
      \textbf{Model} & \textbf{Format} & \textbf{Original} & \textbf{Before faFT} & \textbf{After faFT} \\
      \midrule
      \multirow{3}{*}{FCN} 
        & {\it bfloat16} & 92.3 & 91.6 - 92.0 & 92.1 - 92.2 \\
        & {\it float16}  & 92.3 & 91.5 - 92.1 & 92.1 - 92.3 \\
        & {\it float32}  & 92.3 & 91.0 - 92.1 & 92.0 - 92.3 \\
      \midrule
      \multirow{3}{*}{LeNet} 
        & {\it bfloat16} & 63.8 & 62.0 - 62.3 & 62.6 - 63.5 \\
        & {\it float16}  & 64.0 & 62.0 - 62.2 & 62.2 - 63.4 \\
        & {\it float32}  & 63.9 & 56.2 - 57.4 & 63.0 - 63.4 \\
      \midrule
      \multirow{3}{*}{AlexNet} 
        & {\it bfloat16} & 57.2 & 56.5 - 56.6 & 58.0 - 58.2 \\
        & {\it float16}  & 57.3 & 56.6 - 56.7 & 57.2 - 57.9 \\
        & {\it float32}  & 57.3 & 56.0 - 56.2 & 56.7 - 57.0 \\
      \midrule
      \multirow{3}{*}{VGG16} 
        & {\it bfloat16} & 72.5 & 71.9 - 72.2 & 72.7 - 73.4 \\
        & {\it float16}  & 72.5 & 72.1 - 72.3 & 72.2 - 73.3 \\
        & {\it float32}  & 72.7 & 71.3 - 72.1 & 72.2 - 72.9 \\
      \bottomrule
    \end{tabular}
\end{table}

\section{Experimental Evaluation and Analysis}
\label{sec:evaluation}

In this section, we present comprehensive experimental evaluations of the IScSH and faFT techniques, highlighting their effectiveness in preserving model accuracy in the presence of stuck-at bit faults. Throughout all evaluations, using the IScSh technique on exponent and sign bits obtains the original fault-free accuracy for {\it float32}, and an average relative error increase (REI) less than 0.5\% for {\it float16} and {\it bfloat16}, across all 4 models.

Figures~\ref{fig:fcn-iscsh} and~\ref{fig:vgg16-iscsh} shows results of applying the IScSh technique to mitigate stuck-at faults when using FCN and VGG16 models, respectively. The orange bars show the accuracy impacted by the fault, and the green and blue bars show the accuracy recovered by using fault mitigation. For instance, when using VGG16 on ImageNet Figure~\ref{fig:vgg16-iscsh}, injecting a down link fault into the first PE in the first row (specifically, Stuck Bit 1 at Bit Position 14) causes the model's top-1 accuracy on ImageNet to plummet to 0.1\%. However, after applying IScSh, the accuracy is restored to the original fault-free accuracy of 72.7\%, as shown in the first bar.

While less significant mantissa bits show minimal impact from stuck-at faults ($<1\%$ REI), the most significant mantissa bit positions require fault-aware fine tuning (faFT). Figures~\ref{fig:fcn-faft} and~\ref{fig:vgg16-faft} illustrate faFT's effect for FCN and VGG16 respectivly, with more comprehensive results shown in Table~\ref{tab:faft_table} for all models. For the FCN, faFT used a random subset of 1,000 training examples with a batch size of 64, and 20 epochs with early stopping; for LeNet, 1,000 training examples (batch size 4, 20 epochs, early stopping); for AlexNet, 20,000 examples (batch size 64, 5 epochs, early stopping); and for VGG16, 5000 examples (batch size 40, 5 epochs, early stopping). Overall, the results indicate a substantial recovery in the test accuracy after faFT for all NN architectures across the 3 representations.

The proposed algorithmic approaches, which integrate the faulty behavior instead of bypassing it, are very effective at matching the original fault-free accuracy. We postulate that this is enabled by two types of redundancy: the overprovisioned precision of floating point representation and the overparameterization of neural networks. While theoretically the proposed IScSh operators reducing precision slightly, empirically this had no effect on accuracy, likely due to the overprovisioned precision of the representation. A different kind of redundancy is offered by the NN overparameterization, which gives it sufficient capacity such that ensembles of neurons that learn to recognize the same pattern can still work correctly when only a minority is affected by faults, hence the positive results that we observe in our work when doing fault-aware fine tuning.

\noindent\textbf{Latency Overhead:}
We built a cycle-accurate simulator in C++ to estimate the added latency of the proposed design, across the various test models and input datasets described in Section~\ref{sec:evaluation}. The simulation results are listed in Table~\ref{tab:latency_table}. Across all tests, the proposed algorithmic mitigation techniques induce an average of 17.8\% timing overhead. This overhead is dominated by the nonlinear engine, which operates in parallel with the matrix multiplication engine. While the PE array processes the current tile, the nonlinear engine concurrently rescales the previous tile and prepares the scaling for the next tile. Overhead arises when the nonlinear engine is occupied by other functions (MaxPool in LeNet) or when the scaling operation requires more time than the matrix multiplication as the size of the systolic array increases.

\begin{table}[t]
    \caption{Inference time overhead for different datasets.}
    \label{tab:latency_table}
    \centering
\begin{tabular}{lccc}
\textbf{Systolic Array Size}  & 64$\times$64 & 128$\times$128 & 256$\times$256 \\ \midrule
FCN for MNIST        & 13.1\%       & 14.7\%         & 18.8\%         \\
LeNet for CIFAR-10   & 20.6\%       & 22.0\%         & 22.0\%         \\
AlexNet for ImageNet & 14.0\%       & 17.0\%         & 17.9\%         \\ \bottomrule
\end{tabular}
\vspace{-2em}
\end{table}

\section{Related Work}
\label{sec:relatedwork}

\noindent\textbf{Faults in Systolic Arrays:}
Permanent and transient faults in systolic array-based NN accelerators~\cite{kung1982systolic,jouppi2017datacenter, chen2016eyeriss, genc2019gemmini,li2022tcas,li2022tpds,li:2021:gcnax,wang2022agape,wang2020cure}, including bit-flips, process variation faults, timing errors, and many others can have a negative impact on NN inference accuracy. Stuck-at faults~\cite{patel1998stuck,kajihara1993cost} have emerged as a major challenge in these accelerators. Both permanent and transient stuck-at 0 or 1 faults similarly affect NN output. One study shows that transient faults in higher-order exponent bits cause significant deviations, and  bit flips from 0 to 1 are more severe than from 1 to 0~\cite{LiSC}\cite{ChenZitao}, findings that align with our observations on permanent faults (Section~\ref{sec:characterization}). For detection of stuck-at faults, software and hardware-based built-in-self-tests (BIST)~\cite{BIST1,BIST2,BIST3} are introduced to identify defects. Similar to \cite{caselli2012cooperative, bistnoc1,bistnoc2}, we adopt BIST for the accurate detection at stuck-at faults, which is proven to have 100\% detection coverage for stuck-at permanent faults and can locate the faulty link accurately.


\noindent\textbf{Fault Recovery by Pruning and Training:} 
Recovery from permanent faults using techniques like pruning~\cite{Burel} and fine tuning~\cite{zhang_analyzing_2018} often reduces capacity, particularly in small NNs where all parameters are crucial. Existing fine-tuning methods overlook the impact of the fault during forward and backpropagation. For example, FAP+T~\cite{zhang_analyzing_2018} sets weights of faulty PEs to 0, causing a mismatch between training and inference. Our fault-aware fine tuning (Section~\ref{sec:ft}) instead accounts for faulty weights and activations in gradient calculations and focuses on specific mantissa bits. Compared to FAP+T, our approach has the advantage that it preserves the NN capacity and does not incur any area overhead.

Fault Aware Training (FAT) is a technique for enhancing the reliability of neural networks under hardware faults by incorporating error injection during training~\cite{zahid2020fat}. FAT works by implementing a specialized error injection layer in PyTorch, which simulates hardware-induced errors in activation tensors with a predefined probability, reflecting real-world fault models. During training, these random injected errors condition the network to maintain high accuracy even in the presence of faults. In contrast, the Fault-Aware Fine-Tuning (FaFT) method proposed in this paper does not inject errors at random, instead it fine-tunes the model in the presence of the actual fault detected by BIST, allowing a much more targeted adaptation to the faulty behavior.



Range restriction methods~\cite{ChenZitao} use the range of values observed in the training data to bound values, which allows for some inaccuracy in the values after they are bound. Our IScSh approach is more suitable for critical applications that require precise results, irrespective of the model and dataset.

\noindent\textbf{Hardware-based Solutions:} Hardware mitigation solutions avoid the negative impact of permanent faults by implementing redundant hardware components, including MAC units, PEs, and links~\cite{santos2017applying,takanami2017built,oh2002error,xu2019resilient,wang2019high,li2019squeezing,levine1976special,chang2011design,tsai2011fault,zheng2021adapt,reagen2016minerva,salami2018resilience,intellinoc,wang2015pedestrian,zheng2020versa,takanami1995neural}, {\it inter alia}. However, these techniques inevitably incur significant cost, excessive power consumption, and additional latency.


\section{Conclusion}
\label{sec:conclusion}

We introduced three algorithmic mitigation techniques for a subset of stuck-at faults on links and weight registers in systolic arrays used by NN accelerators: invertible scaling or shifting of activations and parameters, fault-aware fine tuning, and elementary tile operations. Notably, the proposed techniques do not require any hardware modification and integrate the faulty behavior as opposed to bypassing it, offering a more sustainable approach to accelerator reuse.
We implemented the algorithmic mitigation techniques and fault injection methods into a CUDA-accelerated software simulation of systolic arrays (S3A). Extensive experimental evaluations with fully connected and convolutional neural networks trained on MNIST, CIFAR-10, and ImageNet show that the proposed fault-tolerant algorithms match very closely the original fault-free accuracy. 

The PyTorch and CUDA code for the S3A simulator and fault-mitigation methods are made publicly available at \url{https://github.com/yaitalam/s3a}.

This research was partially supported by grant SHF-1901192 from the NSF.


\bibliographystyle{IEEEtranS}
\bibliography{refs}

\begin{thebibliography}{10}
\providecommand{\url}[1]{#1}
\csname url@samestyle\endcsname
\providecommand{\newblock}{\relax}
\providecommand{\bibinfo}[2]{#2}
\providecommand{\BIBentrySTDinterwordspacing}{\spaceskip=0pt\relax}
\providecommand{\BIBentryALTinterwordstretchfactor}{4}
\providecommand{\BIBentryALTinterwordspacing}{\spaceskip=\fontdimen2\font plus
\BIBentryALTinterwordstretchfactor\fontdimen3\font minus \fontdimen4\font\relax}
\providecommand{\BIBforeignlanguage}[2]{{%
\expandafter\ifx\csname l@#1\endcsname\relax
\typeout{** WARNING: IEEEtranS.bst: No hyphenation pattern has been}%
\typeout{** loaded for the language `#1'. Using the pattern for}%
\typeout{** the default language instead.}%
\else
\language=\csname l@#1\endcsname
\fi
#2}}
\providecommand{\BIBdecl}{\relax}
\BIBdecl

\bibitem{reagen2016minerva}
``Minerva: Enabling low-power, highly-accurate deep neural network accelerators,'' in \emph{Proc. of ISCA'16}, 2016.

\bibitem{intellinoc}
``Intellinoc: A holistic design framework for energy-efficient and reliable on-chip communication for manycores,'' in \emph{Proc. of ISCA'19}, 2019.

\bibitem{wang2020cure}
``Cure: A high-performance, low-power, and reliable network-on-chip design using reinforcement learning,'' \emph{IEEE TPDS}, vol.~31, no.~9, pp. 2125--2138, 2020.

\bibitem{zheng2020versa}
``A versatile and flexible chiplet-based system design for heterogeneous manycore architectures,'' in \emph{Proc. of DAC'20}, 2020.

\bibitem{zheng2021adapt}
``Adapt-noc: A flexible network-on-chip design for heterogeneous manycore architectures,'' in \emph{Proc. of HPCA'21}, 2021.

\bibitem{bonderson:talk21}
\BIBentryALTinterwordspacing
``Training in turmoil: Silent data corruption in systems at scale,'' 2021. [Online]. Available: \url{https://marcello.altervista.org/SLM.tttc- events.org/program.html#Keynote1}
\BIBentrySTDinterwordspacing

\bibitem{li2022tcas}
``Ascend: A scalable and energy-efficient deep neural network accelerator with photonic interconnects,'' \emph{IEEE TCAS-I}, vol.~69, no.~7, pp. 2730--2741, 2022.

\bibitem{BIST1}
P.~H. Bardell, W.~H. McAnney, and J.~Savir, \emph{Built-in test for VLSI: pseudorandom techniques}.\hskip 1em plus 0.5em minus 0.4em\relax Wiley-Interscience, 1987.

\bibitem{Burel}
S.~Burel, A.~Evans, and L.~Anghel, ``Mozart: Masking outputs with zeros for architectural robustness and testing of dnn accelerators,'' in \emph{Proceedings of International Symposium on On-Line Testing and Robust System Design (IOLTS)}, 2021, pp. 1--6.

\bibitem{caselli2012cooperative}
N.~Caselli, A.~Strano, D.~Ludovici, and D.~Bertozzi, ``Cooperative built-in self-testing and self-diagnosis of noc bisynchronous channels,'' in \emph{Proceedings of International Symposium on Embedded Multicore SoCs}.\hskip 1em plus 0.5em minus 0.4em\relax IEEE, 2012, pp. 159--166.

\bibitem{chang2011design}
Y.-C. Chang, C.-T. Chiu, S.-Y. Lin, and C.-K. Liu, ``On the design and analysis of fault tolerant noc architecture using spare routers,'' in \emph{Proceedings of Asia and South Pacific Design Automation Conference (ASP-DAC)}, 2011.

\bibitem{chen2016eyeriss}
Y.-H. Chen, T.~Krishna, J.~S. Emer, and V.~Sze, ``Eyeriss: An energy-efficient reconfigurable accelerator for deep convolutional neural networks,'' \emph{IEEE journal of solid-state circuits}, vol.~52, no.~1, pp. 127--138, 2016.

\bibitem{ChenZitao}
Z.~Chen, G.~Li, and K.~Pattabiraman, ``A low-cost fault corrector for deep neural networks through range restriction,'' in \emph{Proceedings of IEEE/IFIP International Conference on Dependable Systems and Networks (DSN)}, 2021, pp. 1--13.

\bibitem{PyTorch}
P.~Contributors, ``Numerical accuracy,'' 2023, \url{https://pytorch.org/docs/stable/notes/numerical_accuracy.html#numerical-accuracy} [Accessed: (04/16/2024)].

\bibitem{dixit_detecting_2022}
\BIBentryALTinterwordspacing
H.~D. Dixit, L.~Boyle, G.~Vunnam, S.~Pendharkar, M.~Beadon, and S.~Sankar, ``Detecting silent data corruptions in the wild,'' Mar. 2022, arXiv:2203.08989 [cs]. [Online]. Available: \url{http://arxiv.org/abs/2203.08989}
\BIBentrySTDinterwordspacing

\bibitem{dosovitskiy_image_2020}
\BIBentryALTinterwordspacing
A.~Dosovitskiy, L.~Beyer, A.~Kolesnikov, D.~Weissenborn, X.~Zhai, T.~Unterthiner, M.~Dehghani, M.~Minderer, G.~Heigold, S.~Gelly, J.~Uszkoreit, and N.~Houlsby, ``\BIBforeignlanguage{en}{An {Image} is {Worth} 16x16 {Words}: {Transformers} for {Image} {Recognition} at {Scale}},'' in \emph{\BIBforeignlanguage{en}{Proceedings of International Conference on Learning Representations (ICLR)}}, 2021. [Online]. Available: \url{https://openreview.net/forum?id=YicbFdNTTy}
\BIBentrySTDinterwordspacing

\bibitem{BIST2}
E.~B. Eichelberger and E.~Lindbloom, ``Random-pattern coverage enhancement and diagnosis for lssd logic self-test,'' \emph{IBM Journal of Research and Development}, vol.~27, no.~3, pp. 265--272, 1983.

\bibitem{genc2019gemmini}
H.~Genc, S.~Kim, A.~Amid, A.~Haj-Ali, V.~Iyer, P.~Prakash, J.~Zhao, D.~Grubb, H.~Liew, H.~Mao, A.~Ou, C.~Schmidt, S.~Steffl, J.~Wright, I.~Stoica, J.~Ragan-Kelley, K.~Asanovic, B.~Nikolic, and Y.~S. Shao, ``Gemmini: Enabling systematic deep-learning architecture evaluation via full-stack integration,'' in \emph{2021 58th ACM/IEEE Design Automation Conference (DAC)}, 2021, pp. 769--774.

\bibitem{grecu2006bist}
C.~Grecu, P.~Pande, A.~Ivanov, and R.~Saleh, ``Bist for network-on-chip interconnect infrastructures,'' in \emph{24th IEEE VLSI Test Symposium}.\hskip 1em plus 0.5em minus 0.4em\relax IEEE, 2006, pp. 6--pp.

\bibitem{he_understanding_2023}
\BIBentryALTinterwordspacing
Y.~He, M.~Hutton, S.~Chan, R.~De~Gruijl, R.~Govindaraju, N.~Patil, and Y.~Li, ``Understanding and {Mitigating} {Hardware} {Failures} in {Deep} {Learning} {Training} {Systems},'' in \emph{Proceedings of ISCA}.\hskip 1em plus 0.5em minus 0.4em\relax New York, NY, USA: Association for Computing Machinery, Jun. 2023, pp. 1--16. [Online]. Available: \url{https://dl.acm.org/doi/10.1145/3579371.3589105}
\BIBentrySTDinterwordspacing

\bibitem{hochschild_cores_2021}
\BIBentryALTinterwordspacing
P.~H. Hochschild, P.~Turner, J.~C. Mogul, R.~Govindaraju, P.~Ranganathan, D.~E. Culler, and A.~Vahdat, ``Cores that don't count,'' in \emph{Proceedings of the {Workshop} on {Hot} {Topics} in {Operating} {Systems}}.\hskip 1em plus 0.5em minus 0.4em\relax Association for Computing Machinery, Jun. 2021, pp. 9--16. [Online]. Available: \url{https://dl.acm.org/doi/10.1145/3458336.3465297}
\BIBentrySTDinterwordspacing

\bibitem{takanami1995neural}
K.~K. I.~Takanami and T.~Watanabe, ``A neural algorithm for reconstructing mesh-connected processor arrays using single-track switches,'' in \emph{Proc. of ICWSI'95}.\hskip 1em plus 0.5em minus 0.4em\relax IEEE, 1995, pp. 101--110.

\bibitem{li:2021:gcnax}
A.~K. J.~Li, A.~Louri and R.~Bunescu, ``Gcnax: A flexible and energy-efficient accelerator for graph convolutional neural networks,'' in \emph{Proc. of HPCA'21}.\hskip 1em plus 0.5em minus 0.4em\relax IEEE, 2021, pp. 775--788.

\bibitem{li2022tpds}
K.~W. J.~Li, H.~Zheng and A.~Louri, ``{SGCNAX}: A scalable graph convolutional neural network accelerator with workload balancing,'' \emph{IEEE TPDS}, vol.~33, no.~11, pp. 2834--2845, 2022.

\bibitem{jawahir_technological_2016}
\BIBentryALTinterwordspacing
I.~S. Jawahir and R.~Bradley, ``Technological {Elements} of {Circular} {Economy} and the {Principles} of {6R}-{Based} {Closed}-loop {Material} {Flow} in {Sustainable} {Manufacturing},'' \emph{Procedia CIRP}, vol.~40, pp. 103--108, Jan. 2016. [Online]. Available: \url{https://www.sciencedirect.com/science/article/pii/S2212827116000822}
\BIBentrySTDinterwordspacing

\bibitem{jouppi2017datacenter}
N.~P. Jouppi, C.~Young, N.~Patil, D.~Patterson, G.~Agrawal, R.~Bajwa, S.~Bates, S.~Bhatia, N.~Boden, A.~Borchers \emph{et~al.}, ``In-datacenter performance analysis of a tensor processing unit,'' in \emph{Proceedings of ISCA)}, 2017, pp. 1--12.

\bibitem{wang2019high}
A.~K. K.~Wang, A.~Louri and R.~Bunescu, ``High-performance, energy-efficient, fault-tolerant network-on-chip design using reinforcement learning,'' in \emph{Proc. of DATE'19}, 2019, pp. 1166--1171.

\bibitem{kajihara1993cost}
S.~Kajihara, I.~Pomeranz, K.~Kinoshita, and S.~M. Reddy, ``Cost-effective generation of minimal test sets for stuck-at faults in combinational logic circuits,'' in \emph{Proceedings of the 30th International Design Automation Conference}, 1993, pp. 102--106.

\bibitem{kalamkar2019study}
D.~Kalamkar, D.~Mudigere, N.~Mellempudi, D.~Das, K.~Banerjee, S.~Avancha, D.~T. Vooturi, N.~Jammalamadaka, J.~Huang, H.~Yuen, J.~Yang, J.~Park, A.~Heinecke, E.~Georganas, S.~Srinivasan, A.~Kundu, M.~Smelyanskiy, B.~Kaul, and P.~Dubey, ``A study of bfloat16 for deep learning training,'' 2019.

\bibitem{krizhevsky_imagenet_2012}
A.~Krizhevsky, I.~Sutskever, and G.~E. Hinton, ``{ImageNet} {Classification} with {Deep} {Convolutional} {Neural} {Networks},'' in \emph{NeurIPS}, vol.~25.\hskip 1em plus 0.5em minus 0.4em\relax Curran Associates, Inc., 2012.

\bibitem{Kung}
Kung, ``Why systolic architectures?'' \emph{Computer}, vol.~15, no.~1, pp. 37--46, 1982.

\bibitem{kung1982systolic}
H.-T. Kung, ``Why systolic architectures?'' \emph{Computer}, 1982.

\bibitem{BIST3}
J.~J. LeBlanc, ``Locst: A built-in self-test technique,'' \emph{IEEE Design \& Test of Computers}, vol.~1, no.~4, pp. 45--52, 1984.

\bibitem{lecun_gradient-based_1998}
\BIBentryALTinterwordspacing
Y.~Lecun, L.~Bottou, Y.~Bengio, and P.~Haffner, ``Gradient-based learning applied to document recognition,'' \emph{Proceedings of the IEEE}, vol.~86, no.~11, pp. 2278--2324, Nov. 1998, conference Name: Proceedings of the IEEE. [Online]. Available: \url{https://ieeexplore.ieee.org/document/726791}
\BIBentrySTDinterwordspacing

\bibitem{levine1976special}
L.~Levine and W.~Meyers, ``Special feature: Semiconductor memory reliability with error detecting and correcting codes,'' \emph{Computer}, vol.~9, no.~10, pp. 43--50, 1976.

\bibitem{LiSC}
\BIBentryALTinterwordspacing
G.~Li, S.~K.~S. Hari, M.~Sullivan, T.~Tsai, K.~Pattabiraman, J.~Emer, and S.~W. Keckler, ``Understanding error propagation in deep learning neural network (dnn) accelerators and applications,'' in \emph{Proceedings of SC}, ser. SC '17.\hskip 1em plus 0.5em minus 0.4em\relax New York, NY, USA: Association for Computing Machinery, 2017. [Online]. Available: \url{https://doi.org/10.1145/3126908.3126964}
\BIBentrySTDinterwordspacing

\bibitem{li2019squeezing}
L.~Li, D.~Xu, K.~Xing, C.~Liu, Y.~Wang, H.~Li, and X.~Li, ``Squeezing the last mhz for cnn acceleration on {FPGA}s,'' in \emph{Proc. of ITC-Asia'19}.\hskip 1em plus 0.5em minus 0.4em\relax IEEE, 2019, pp. 151--156.

\bibitem{bistnoc1}
S.-Y. Lin, C.-C. Hsu, and A.-Y. Wu, ``A scalable built-in self-test/self-diagnosis architecture for 2d-mesh based chip multiprocessor systems,'' in \emph{2009 IEEE International Symposium on Circuits and Systems}.\hskip 1em plus 0.5em minus 0.4em\relax IEEE, 2009, pp. 2317--2320.

\bibitem{mishra2023artificial}
A.~Mishra, J.~Cha, H.~Park, and S.~Kim, \emph{Artificial Intelligence and Hardware Accelerators}.\hskip 1em plus 0.5em minus 0.4em\relax Springer, 2023.

\bibitem{oh2002error}
N.~Oh, P.~Shirvani, and E.~McCluskey, ``Error detection by duplicated instructions in super-scalar processors,'' \emph{IEEE Trans. Reliab.}, 2002.

\bibitem{painegpu:iclr14}
T.~Paine, H.~Jin, J.~Yang, Z.~Lin, and T.~Huang, ``Gpu asynchronous stochastic gradient descent to speed up neural network training, corr abs/1312.6186,'' in \emph{Proceedings of the 2nd International Conference on Learning Representations (ICLR 2014)}, April 2014.

\bibitem{patel1998stuck}
J.~H. Patel, ``Stuck-at fault: a fault model for the next millennium,'' in \emph{Proceedings of International Test Conference}.\hskip 1em plus 0.5em minus 0.4em\relax IEEE, 1998, p. 1166.

\bibitem{bistnoc2}
K.~Peters{\'e}n and J.~{\"O}berg, ``Utilizing noc switches as bist structures in 2d-mesh network-on-chips,'' in \emph{DATE-2006. Munich, Germany. 6-10 March 2006}, 2006.

\bibitem{salami2018resilience}
B.~Salami, O.~S. Unsal, and A.~C. Kestelman, ``On the resilience of rtl nn accelerators: Fault characterization and mitigation,'' in \emph{Proc. of SBAC-PAD'18}.\hskip 1em plus 0.5em minus 0.4em\relax IEEE, 2018.

\bibitem{santos2017applying}
A.~Santos, L.~Antunes~Tambara, F.~Benevenuti, J.~Tonfat, and F.~Kastensmidt, ``Applying tmr in hardware accelerators generated by high-level synthesis design flow for mitigating multiple bit upsets in sram-based {FPGA}s,'' in \emph{Proc. of ARC'17}.\hskip 1em plus 0.5em minus 0.4em\relax Springer, 2017.

\bibitem{simonyan_very_2015}
\BIBentryALTinterwordspacing
K.~Simonyan and A.~Zisserman, ``Very {Deep} {Convolutional} {Networks} for {Large}-{Scale} {Image} {Recognition},'' in \emph{International Conference of Learning Representations (ICLR)}, Apr. 2015, arXiv:1409.1556. [Online]. Available: \url{http://arxiv.org/abs/1409.1556}
\BIBentrySTDinterwordspacing

\bibitem{takanami2017built}
I.~Takanami and M.~Fukushi, ``A built-in circuit for self-repairing mesh-connected processor arrays with spares on diagonal,'' in \emph{Proc. of PRDC'17}.\hskip 1em plus 0.5em minus 0.4em\relax IEEE, 2017, pp. 110--117.

\bibitem{tsai2011fault}
W.-C. Tsai, D.-Y. Zheng, S.-J. Chen, and Y.-H. Hu, ``A fault-tolerant noc scheme using bidirectional channel,'' in \emph{Proc. of DAC'11}, 2011, pp. 918--923.

\bibitem{vaswani_attention_2017}
\BIBentryALTinterwordspacing
A.~Vaswani, N.~Shazeer, N.~Parmar, J.~Uszkoreit, L.~Jones, A.~N. Gomez, L.~Kaiser, and I.~Polosukhin, ``Attention is {All} you {Need},'' in \emph{Advances in {Neural} {Information} {Processing} {Systems}}, vol.~30.\hskip 1em plus 0.5em minus 0.4em\relax Curran Associates, Inc., 2017. [Online]. Available: \url{https://papers.nips.cc/paper_files/paper/2017/hash/3f5ee243547dee91fbd053c1c4a845aa-Abstract.html}
\BIBentrySTDinterwordspacing

\bibitem{wang2022agape}
K.~Wang, H.~Zheng, Y.~Li, J.~Li, and A.~Louri, ``{AGAPE}: anomaly detection with generative adversarial network for improved performance, energy, and security in manycore systems,'' in \emph{Proc. of DATE'22}.\hskip 1em plus 0.5em minus 0.4em\relax IEEE, 2022.

\bibitem{wang2015pedestrian}
R.~Wang and Z.~Xu, ``A pedestrian and vehicle rapid identification model based on convolutional neural network,'' in \emph{Proc. of ICIMCS'15}, 2015, pp. 1--4.

\bibitem{xu2019resilient}
D.~Xu, K.~Xing, C.~Liu, Y.~Wang, Y.~Dai, L.~Cheng, H.~Li, and L.~Zhang, ``Resilient neural network training for accelerators with computing errors,'' in \emph{Proc. of ASAP}.\hskip 1em plus 0.5em minus 0.4em\relax IEEE, 2019.

\bibitem{zahid2020fat}
U.~Zahid, G.~Gambardella, N.~J. Fraser, M.~Blott, and K.~Vissers, ``Fat: Training neural networks for reliable inference under hardware faults,'' in \emph{2020 IEEE International Test Conference (ITC)}.\hskip 1em plus 0.5em minus 0.4em\relax IEEE, 2020, pp. 1--10.

\bibitem{zhang_analyzing_2018}
J.~J. Zhang, T.~Gu, K.~Basu, and S.~Garg, ``Analyzing and mitigating the impact of permanent faults on a systolic array based neural network accelerator,'' in \emph{VTS)}, Apr. 2018, pp. 1--6, iSSN: 2375-1053.

\bibitem{zhou2018nvdla}
G.~Zhou, J.~Zhou, and H.~Lin, ``Research on {NVIDIA} deep learning accelerator,'' in \emph{2018 12th IEEE International Conference on Anti-counterfeiting, Security, and Identification (ASID)}.\hskip 1em plus 0.5em minus 0.4em\relax IEEE, 2018, pp. 192--195.

\bibitem{zinkevich_parallelized_2010}
\BIBentryALTinterwordspacing
M.~Zinkevich, M.~Weimer, L.~Li, and A.~Smola, ``Parallelized {Stochastic} {Gradient} {Descent},'' in \emph{Advances in {Neural} {Information} {Processing} {Systems}}, vol.~23.\hskip 1em plus 0.5em minus 0.4em\relax Curran Associates, Inc., 2010. [Online]. Available: \url{https://papers.nips.cc/paper_files/paper/2010/hash/abea47ba24142ed16b7d8fbf2c740e0d-Abstract.html}
\BIBentrySTDinterwordspacing

\end{thebibliography}

\end{document}